\documentclass[a4paper,fleqn]{cas-dc}

\usepackage[numbers]{natbib}
\usepackage{amsmath}
\usepackage{algorithmic}
\usepackage{graphicx}
\usepackage{textcomp}
\usepackage{xcolor}
\usepackage{bm}
\usepackage{amstext}
\usepackage{multirow}
\usepackage{times}
\usepackage{epsfig}
\usepackage{amsfonts}
\usepackage{amssymb}
\usepackage{color}
\usepackage{hyperref}

\def\tsc#1{\csdef{#1}{\textsc{\lowercase{#1}}\xspace}}
\tsc{WGM}
\tsc{QE}
\tsc{EP}
\tsc{PMS}
\tsc{BEC}
\tsc{DE}

\begin{document}
\shorttitle{}
\shortauthors{Zhenxing Zhang et~al.}

\title [mode = title]{OptGAN: Optimizing and Interpreting the Latent Space of the Conditional Text-to-Image GANs} 

\author{Zhenxing Zhang}
\cormark[1]
\ead{z.zhang@rug.nl}

\credit{Conceptualization, Data curation, Formal analysis, Investigation, Methodology, Software, Validation, Visualization, Writing - original draft}

\address{Bernoulli Institute, Faculty of Science and Engineering, University of Groningen, Groningen 9747, The Netherlands}

\author{Lambert Schomaker}
\ead{l.r.b.schomaker@rug.nl}

\credit{Conceptualization, Formal analysis, Methodology, Funding acquisition, Resources, Project administration, Supervision, Writing - review $\&$ editing}

\cortext[cor1]{Corresponding author}
\begin{abstract}
Text-to-image generation intends to automatically produce a photo-realistic image, conditioned on a textual description. It can be potentially employed in the field of art creation, data augmentation, photo-editing, etc. Although many efforts have been dedicated to this task, it remains particularly challenging to generate believable, natural scenes. To facilitate the real-world applications of text-to-image synthesis, we focus on studying the following three issues: 1) How to ensure that generated samples are believable, realistic or natural? 2) How to exploit the latent space of the generator to edit a synthesized image? 3) How to improve the explainability of a text-to-image generation framework? In this work, we constructed two novel data sets (i.e., the {\em Good \& Bad} bird and face data sets) consisting of successful as well as unsuccessful generated samples, according to strict criteria. To effectively and efficiently acquire high-quality images by increasing the probability of generating ${Good}$ latent codes, we use a dedicated Good/Bad classifier for generated images. It is based on a pre-trained front end and fine-tuned on the basis of the proposed {\em Good \& Bad} data set. After that, we present a novel algorithm which identifies semantically-understandable directions in the latent space of a conditional text-to-image GAN architecture by performing independent component analysis on the pre-trained weight values of the generator. Furthermore, we develop a background-flattening loss (BFL), to improve the background appearance in the edited image. Subsequently, we introduce linear interpolation analysis between pairs of keywords. This is extended into a similar triangular `linguistic' interpolation in order to take a deep look into what a text-to-image synthesis model has learned within the linguistic embeddings. Experimental results on the recent DiverGAN generator pre-trained on three benchmark data sets demonstrate that our classifier achieves a better than 98\% accuracy in predicting Good/Bad classes for synthetic samples and our proposed approach is able to derive various interpretable semantic properties for a conditional text-to-image GAN model, confirming the effectiveness of our presented techniques. Our data set is available at https://zenodo.org/record/6283798$\#$.YhkN$\_$ujMI2w. 
\end{abstract}


\maketitle

\section{Introduction}
The task of text-to-image synthesis aims at automatically generating high-quality and semantically-consistent images, given  natural-language descriptions. It has recently gathered increasing interest from researchers due to its numerous potential applications, e.g., data augmentation for training image classifiers, photo editing according to textual descriptions, the education of young children, etc. With the advances in the generative adversarial network (GAN) and the conditional generative adversarial network (cGAN)  \cite{mirza2014conditional}, text-to-image generation has achieved promising progress in both image quality and semantic consistency. Nevertheless, it remains extremely challenging to coerce a conditional text-to-image GAN model to generate, with high probability, believable and natural images.  

One particular disadvantage of synthetic image-generation algorithms is that the performance evaluation is more difficult than is the case in classification problems where a `hard' accuracy can be computed. In case of the cGAN this issue is most clearly present for end users: \textit{How to ensure that generated images are believable, realistic or natural?} In current literature, the good examples are often cherry picked while occasionally also the less successful samples are shown. However, for actual use in data augmentation or in artistic applications, one would like to guarantee that generated images are good, i.e., of a sufficiently believable natural quality. Given the high dimensionality of latent codes, there is a very high prior probability of non-successful patterns to be generated for a given input noise probe. How to construct a random latent-code generator with an increased probability of drawing successful samples? After the generator/discriminator pair has done its best effort, apparently additional constraints are necessary.
\begin{figure*}
  \begin{minipage}[b]{1.0\linewidth}
  \centerline{\includegraphics[width=180mm]{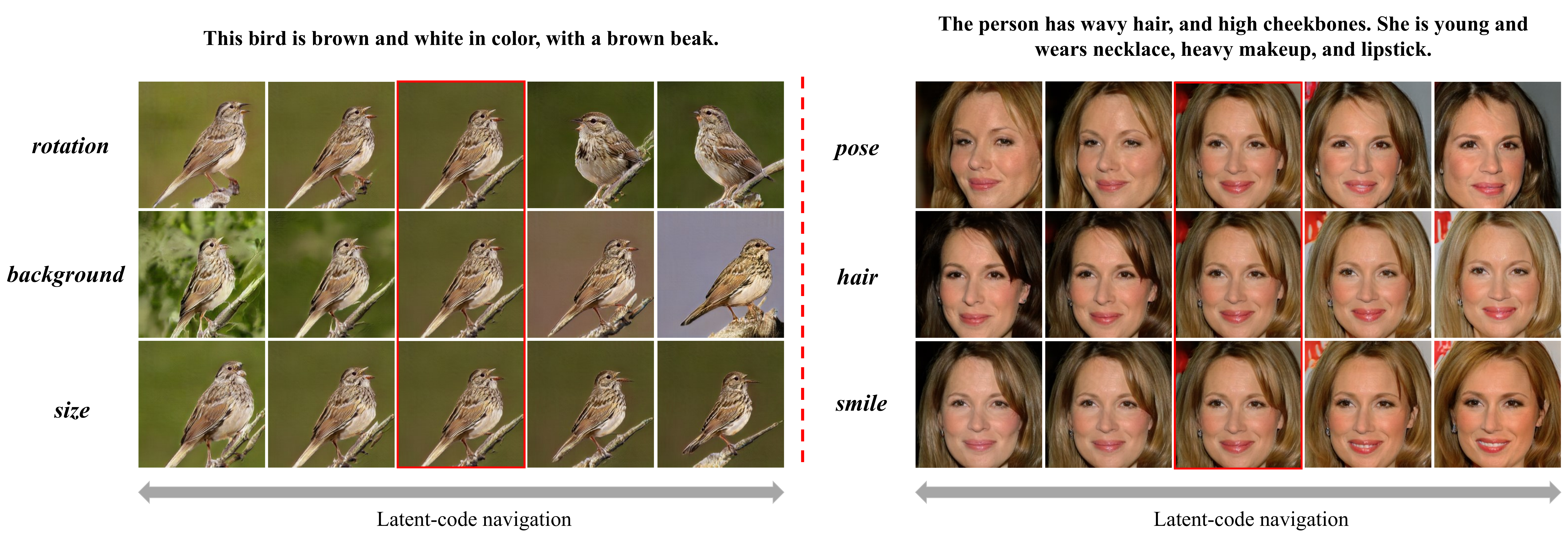}}
  \end{minipage}
  \caption{Interpretable latent-space directions identified in DiverGAN \cite{zhang2021divergan} that was pre-trained on the CUB bird \cite{wah2011caltech} (\textbf{left} side) and Multi-Modal CelebA-HQ \cite{xia2021tedigan} (\textbf{right} side) data sets. For each set of pictures, the middle column is the original image based on a ${Good}$ latent code, while the samples on the left and right of it are the output by freezing the textual description and moving the latent vector backward and forward from the center, over the axis discovered by our proposed algorithm.}
  \vspace{-0.1in}
  \label{fig01} 
\end{figure*}

Here, we intend to train a classifier to accurately distinguish successful synthesized samples from unsuccessful generated pictures after training a text-to-image generation framework. This is based on the assumption that there is a non-linear boundary separating high-resolution images from inadequate samples in the fake image space. To this end, we created a {\em Good \& Bad} data set, both for a bird and a face-image collection (shown in Fig.~\ref{fig03}), which consists of a large number of realistic as well as implausible samples synthesized by the recent DiverGAN \cite{zhang2021divergan} that was pretrained on the CUB bird \cite{wah2011caltech} data set and the Multi-Modal CelebA-HQ data set \cite{xia2021tedigan}, respectively. We choose these samples by following strict principles in order to ensure the quality of the selected images. To acquire a superior classifier, we train the CNN model (e.g., ResNet \cite{he2016deep}) from the pre-trained weights on our {\em Good \& Bad} data set. We expect that the well-trained network can correctly predict the quality class of synthesized images.
Therefore, we are able to effectively and efficiently derive photo-realistic images from the synthesized samples while obtaining corresponding ${Good}$ latent vectors. More importantly, the discovery of ${Good}$ latent codes provides a strong basis for further research, such as data augmentation and latent-space manipulation. 

Latent vectors contributing to diversity play a significant role in the image-generation process. Recent works \cite{shen2021closed, harkonen2020ganspace, shen2020interpreting} reveal that there exists a wide range of meaningful semantic factors in the latent space of a GAN, such as facial attributes and head poses for face synthesis \cite{shen2020interpreting} and layout for scene generation \cite{yang2021semantic}. These semantically-understandable control directions can be utilized for disentangled image editing, like semantic face editing \cite{shen2020interpreting} and scene manipulation \cite{yang2021semantic}. By moving the latent code of a synthetic sample towards and backwards the direction, we are able to vary the desired attribute while keeping other image contents unchanged. That is to say, given a successful latent code, we can derive a wealth of similar but semantically-diverse pleasing images via latent-space navigation. To better facilitate the application of text-to-image synthesis, we need to address the question: \textit{How to identify useful control directions in the latent space of a conditional text-to-image GAN model?} While current approaches mainly focus on studying the latent space of a GAN, there still is a lack of understanding of the relationship between the latent space of a cGAN and the, explainable semantic space in which a synthetic sample is embedded. 

In this paper, we present a novel algorithm to capture the interpretable latent-space semantic properties for a text-to-image synthesis model. Considering the fact that identified directions denote different semantic factors of the edited object (e.g., pose and smile for the face model), we argue that these vectors should be fully independent rather than just uncorrelated. Based on recent studies \cite{shen2021closed, harkonen2020ganspace}, we assume that the pre-trained weights of a conditional text-to-image GAN architecture contain a set of useful directions. In fact, the initial linear layer projects the latent vector to the visual feature map, where a latent space is transformed into another space and ultimately into an output image. To acquire both independent and orthogonal components, we introduce the independent component analysis (ICA) algorithm under an additional orthogonality constraint \cite{ablin2018faster} to investigate the pre-trained weight matrix of the first dense layer. In addition, we mathematically show that Semantic Factorization (SeFa) \cite{shen2021closed}, GANSpace \cite{harkonen2020ganspace} and regular PCA \cite{wold1987principal} typically achieve almost the identical results when sampling enough data for GANSpace. Furthermore, we develop a {\em Background-Flattening Loss} (BFL), to improve the background appearance in the edited sample. Multiple interesting latent-space directions found by our presented algorithm are visualized in Fig. \ref{fig01}.

We expect that our proposed semantic-discovery method can provide valuable insight into the correlation between latent vectors and image variations. However, it remains particularly difficult to explain what a conditional text-to-image GAN model has learned within the text space. \textit{How to understand the relation between the textual (linguistic) probes and the generated image factors?} This constitutes the last research topic of the current contribution. To alleviate the problem, we qualitatively analyze the roles played by the linguistic embeddings in the generated-image semantic space through linear interpolation analysis between pairs of keywords. We show that although semantic properties contained in the picture change continuously in the latent space, the appearance of the image does not always vary smoothly along with the contrasting word embeddings. In addition, we extend a pairwise linear interpolation to a triangular interpolation for simultaneously investigating three keywords in the give textual description.

The recent DiverGAN \cite{zhang2021divergan} has the ability to adopt a generator/discriminator pair to synthesize diverse and high-quality samples, given a textual description and different injected noise on the latent vector. We therefore carry out a serious of experiments on the DiverGAN generator that was trained on three popular text-to-image data sets (i.e., the CUB bird \cite{wah2011caltech}, MS COCO \cite{lin2014microsoft} and Multi-Modal CelebA-HQ \cite{xia2021tedigan} data sets).
The experimental results in the current study represent an improvement in performance and explainability in the analyzed algorithm \cite{zhang2021divergan}. Meanwhile, our well-trained classifier achieves impressive classification accuracy (bird: 98.09$\%$ and face: 99.16$\%$) on the {\em Good \& Bad} data set and our proposed semantic-discovery algorithm can lead to a more precise control over the latent space of the DiverGAN model, which validate the effectiveness of our presented methods. 
The contributions of this work can be summarized as follows:

$\bullet$ We construct two new {\em Good \&
Bad} data sets to study how to ensure that generated images are believable while training two corresponding classifiers to separate successful generated images from unsuccessful synthetic samples.

$\bullet$ We introduce the ICA algorithm to identify meaningful attributes in the latent space of a conditional text-to-image GAN model. Simultaneously, we analyze the correspondences between SeFa, GANSpace and regular PCA.

$\bullet$ We introduce linear interpolation analysis between pairs of contrastive keywords and a similar triangular `linguistic' interpolation for an improved explainability of a text-to-image generation architecture. 

The remainder of the paper is organized as follows. We introduce the related works in Section~\ref{rw}. Section~\ref{preliminary} briefly depicts the single-stage text-to-image framework and the corresponding latent space. In Section~\ref{pa}, we describe our OptGAN approach in detail. The experimental results are presented in Section~\ref{er} and Section 6 draws the conclusions.

\section{Related works}
\label{rw}
In this section, we depict the research fields associated with our work, i.e., a GAN, cGAN-based text-to-image generation and latent-space manipulation. 
\subsection{Generative adversarial network (GAN)}
A GAN first presented by Goodfellow et al. \cite{goodfellow2014generative} builds a basic model for synthetic tasks via adversarial training, consisting of a generator and a discriminator. A GAN has achieved state-of-the-art performance in a variety of applications including text-to-image synthesis \cite{gao2021lightweight}, person image generation \cite{shi2022loss}, face photo-sketch synthesis \cite{yan2021isgan}, image inpainting \cite{zhang2021gan}, image de-raining \cite{yang2022rain}, etc, since it is capable of producing photo-realistic images. 

The initial generator network of a GAN mainly comprises multi-layer perceptrons and rectifier linear activations, while the discriminator net utilizes maxout network \cite{goodfellow2013maxout}. This type of architecture shows competitive samples with other generative models on simple image datasets, such as MNIST \cite{lecun1998gradient}. Moreover, researchers explore different structures of a GAN in order to further improve image quality. Denton et al. \cite{denton2015deep} designed a Laplacian pyramid framework of an adversarial network 
namely LAPGAN that produces plausible results in a coarse-to-fine manner. Radford et al. \cite{radford2015unsupervised} introduced a deep convolutional GAN (DCGAN) integrating convolutional layers and Batch Normalization (BN) \cite{ioffe2015batch} into both a generator and a discriminator. Mirza et al. \cite{mirza2014conditional} proposed a cGAN by imposing conditional constraints (e.g., class labels, text descriptions and low-resolution images) on both a generator and a discriminator to obtain specific samples. Recently, several models with a high-computational cost are introduced to yield visually plausible pictures. Zhang et al. \cite{zhang2019self} presented SAGAN which applies the self-attention mechanism to effectively capture the semantic affinities between widely separated image regions. Brock et al. \cite{brock2018large} developed a large-scale architecture based on SAGAN while deploying orthogonal regularization to the generator, obtaining excellent performance on image diversity. Karras et al. \cite{karras2019style} proposed a novel generator framework named StyleGAN where adaptive instance normalization is utilized to control the generator. This paper focuses on studying a conditional text-to-image GAN model.

\subsection{cGAN in text-to-image generation}
Owing to the success of a GAN on image quality, the task of text-to-image synthesis has achieved significant advances over the past few years. Existing approaches for text-to-image generation can be roughly cast into two categories: 1) multi-stage models and 2) single-stage methods.

\textbf{Multi-stage models.} Zhang et al. \cite{zhang2017stackgan, zhang2018stackgan++} introduced a multi-stage architecture called StackGAN, in which each stage comprises a generator and a discriminator, and the generator of the next stage receives the result of the previous stage as the input. Xu et al. \cite{xu2018attngan} proposed AttnGAN inserting a spatial attention module into the multi-stage framework to bridge the semantic gap between the words in a textual description and the related image subregions. Qiao et al. \cite{qiao2019mirrorgan} presented MirrorGAN where an image-to-text model is leveraged to guarantee the semantic consistency between natural-language descriptions and visual contents. Zhu et al. \cite{zhu2019dm} designed DMGAN which introduce a dynamic-memory module to produce high-quality samples in the initial stage. OP-GAN presented by Hinz et al. \cite{hinz2019semantic}   explicitly modeled the objects of an image while developing a new evaluation metric termed as semantic object accuracy.

\textbf{Single-stage methods.} Reed et al. \cite{Radford2016UnsupervisedRL} were the first to attempt to employ the cGAN to synthesize specific images based on the given text descriptions. Tao et al. \cite{tao2020df} proposed DFGAN where a matching-aware zero-centered gradient penalty loss is introduced to help stabilize the training of the conditional text-to-image GAN model. Zhang et al. \cite{zhang2021dtgan} designed DTGAN by utilizing spatial and channel attention modules and the conditional normalization to yield photo-realistic samples with a generator/discriminator pair. Zhang et al. \cite{zhang2021cross} developed XMC-GAN which studied contrastive learning in the context of text-to-image generation while producing visually plausible images via a simple single-stage framework. Zhang et al. \cite{zhang2021divergan} presented an efficient and effective single-stage framework called DiverGAN which is capable of generating diverse, plausible and semantically-consistent images according to a natural-language description. Note that we adopt the DiverGAN generator to perform comprehensive experiments due to its superior performance on image quality and diversity.

\subsection{Study on the Latent Space of a GAN}
Recent studies \cite{shen2020interpreting, harkonen2020ganspace, shen2021closed} on a GAN reveal that a latent space possess a range of semantically-understandable information (e.g., pose and smile for the face data set), which plays a vital role in detangled sample manipulation. We are able to realistically edit the generated image by moving its latent vector towards the direction corresponding to the desired attribute. Several methods have been proposed to capture interpretable semantic factors and mainly fall into two types: 1) unsupervised models and 2) supervised approaches. 

\textbf{Supervised latent-space manipulation.} Shen et al. \cite{shen2020interpreting} developed a framework termed as InterfaceGAN where labeled samples (e.g., gender and age) are utilized to train a linear Support Vector Machine (SVM) and the acquired SVM boundaries lead to the meaningful manipulation of the facial attributes.  Goetschalckx et al. \cite{goetschalckx2019ganalyze} proposed GANalyze applying an accessor module to optimize the training process while learning the latent-space directions as the desired cognitive semantics.

\textbf{Unsupervised latent-space manipulation.} Voynov et al. \cite{voynov2020unsupervised} introduced a matrix and a classifier to identify interpretable latent-space directions in an unsupervised fashion. Jahanian et al. \cite{jahanian2019steerability} studied the attributes concerning color transformations and camera movements by operating source pictures. Härkönen et al. \cite{harkonen2020ganspace} designed a novel pipeline named GANSpace, which performed PCA \cite{wold1987principal} on a series of collected latent vectors and employed obtained principal components as the meaningful directions in the latent space. Peebles et al. \cite{wang2021hijack} presented the Hessian Penalty, a regularization term for the unsupervised discovery of useful semantic factors. Wang et al. \cite{wang2021hijack} developed Hijack-GAN introducing an iterative scheme to control the image-generation process. Shen et al. \cite{shen2021closed} proposed Semantic Factorization (SeFa) which directly decomposed the weight matrix of a well-trained GAN model for semantic image editing. Our work aims to identify controllable directions in the latent space of a conditional text-to-image GAN model.
\section{Preliminary}
\label{preliminary}
In this section, we briefly describe the single-stage text-to-image synthesis architecture and the corresponding latent space to help understand the issues we attempt to address. 

\begin{figure}[t]
  \begin{minipage}[b]{1.0\linewidth}
  \centerline{\includegraphics[width=85mm]{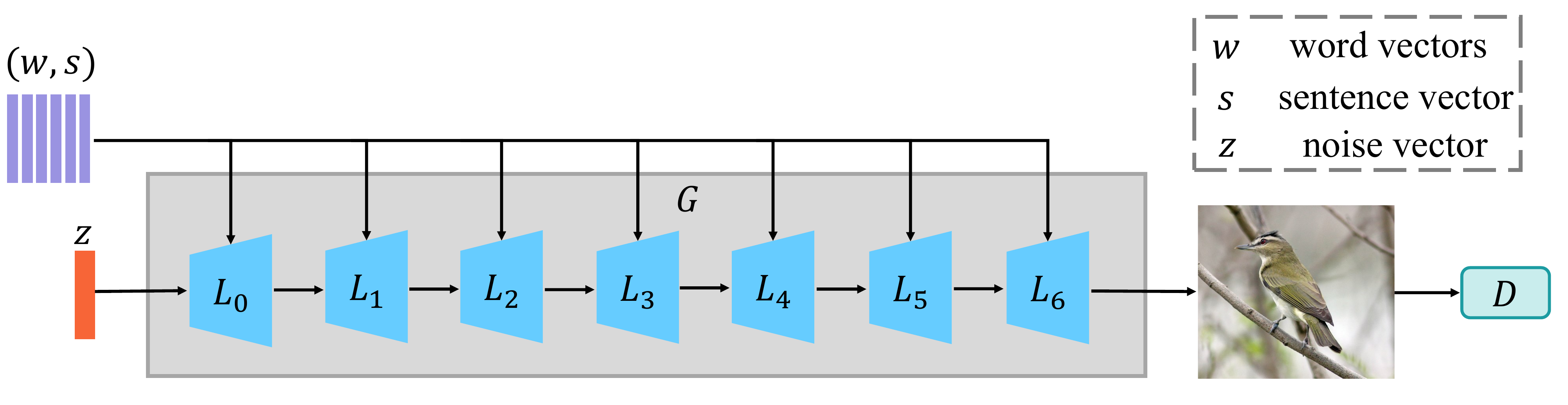}}
  \end{minipage}
  \caption{A simplified single-stage text-to-image generation architecture consisting of a generator ${G}$ and a discriminator ${D}$. The input of the generator is a random latent code ${z}$ and the word/sentence embeddings ${(w, s)}$, and the output is a synthetic sample.}
  \vspace{-0.1in}
  \label{fig02} 
\end{figure}

\textbf{Single-stage pipeline.} The single-stage text-to-image generation framework (illustrated in Fig. \ref{fig02}) is composed of a generator network and a discriminator net, which are perceived as playing a minmax zero-sum game. Let $S=\{ (C_{i}, I_{i})\}_{i=1}^{N}$ denote a collection of $N$ text-image pairs for training, where $I_{i}$ is a picture and $C_{i}=(c_{i}^{1}, c_{i}^{2}, ..., c_{i}^{K})$ comprises $K$ textual descriptions. Word-embedding vectors $w$ and a sentence-embedding vector $s$ are commonly acquired by applying a bidirectional Long Short-Term Memory
(LSTM) network \cite{schuster1997bidirectional} on a natural-language description $c_{i}$ randomly picked from $C_{i}$. After that, the generator $G(z, (w, s))$ is trained to produce a perceptually-realistic and semantically-related image $\hat{I}_{i}$ according to a latent code $z$ randomly sampled from a frozen distribution and word/sentence embedding vectors $(w, s)$. To be specific, $G(z, (w, s))$ consists of multiple layers where the first layer $F_{0}$ maps a latent code into a feature map and intermediate blocks typically leverage modulation modules (e.g., attention models \cite{zhang2021dtgan, zhang2021divergan}) to reinforce the visual feature map to ensure image quality and semantic consistency. The last layer $G_{c}$ transforms the feature map into the ultimate sample. Mathematically,
\begin{align}
&h_{0}=F_{0}(z) \\
&h_{1}=B_{1}(h_{0},(w, s)) \\
&h_{i}=B_{i}(h_{i-1}\uparrow,(w, s))  \quad for \quad i=2,3,...,7 \\
&\hat{I}=G_{c}(h_{7}) 
\end{align}
where $F_{0}$ denotes a fully-connected layer and $B_{i}$ is a modulation block that facilitates the feature map with textual features.

Compared with $G(z, (w, s))$, the discriminator of the single-stage pipeline aims at distinguishing the real text-image pair $(c_{i}, I_{i})$ from the fake text-image pair $(c_{i}, \hat{I}_{i})$.

\textbf{Latent-space analysis.} For a pre-trained and fixed generator $G(z, (w, s))$, the quality of the generated sample depends on the random latent code $z$, word embeddings $w$ and the corresponding sentence vector $s$. Consequently, the output of the network only relies on $z$ when determining the input text description. It implicitly means that if we ignore the linguistic space of the conditional input-text probes, $G(z, (w, s))$ can be regarded as a deterministic function $G$: $\mathcal{Z} \rightarrow \mathcal{X}$. Here, $\mathcal{Z}$ represents the latent space, in which the latent code $z \in R^{l}$ is commonly sampled from a l-dimension Gaussian distribution. $\mathcal{X}$ denotes the synthetic image space including visually realistic samples as well as implausible generated pictures. Moreover, the map from $\mathcal{Z}$ to $\mathcal{X}$ is not surjective \cite{li2021tackling}. Accordingly, even a superior text-to-image generation generator fails to ensure the quality of a synthesized sample, given random latent vectors. In order to promote the applicability of text-to-image generation in practice, this paper intends to optimize the latent space of a conditional text-to-image GAN model to effectively avoid unsuccessful synthetic samples while automatically obtaining high-quality images.

\textbf{Latent-space manipulation.} It has been widely observed that the latent space of a GAN incorporates certain semantic information, like pose and size for the CUB bird data set.
Suppose we have a ${Good}$ latent code $z_{g}$ that contributes to a successful generated sample, and a well-trained generator $G(z, (w, s))$ that can yield dissimilar and semantically consistent pictures according to different textual descriptions and injected noise, we target to manipulate the semantic factor of the successful synthesized sample via latent-space navigation. To this end, we need to first identify a series of semantically-interpretable latent-space directions $N=(n_{1}, n_{2},\cdots, n_{k})$, where $ n_{i} \in R^{l}$ 
for all $i \in {1,2,\cdots,k}$. 
Then, the attribute of the high-quality sample generated by $z_{g}$ can be varied by editing $z_{g}$ with $z_{ge}=z_{g} +\alpha n$, where $\alpha$ denotes the manipulation intensity and $n \in R^{l}$ is the direction corresponding to the desired property.
\section{Proposed methodology}
\label{pa}
In this section, we elaborate on the proposed procedure automatically finding successful synthetic samples from generated images while acquiring corresponding  ${Good}$ latent codes. After that, based on a ${Good}$ latent vector, we describe the independent component analysis method that identifies meaningful latent-space directions for a conditional text-to-image GAN model. Subsequently, we introduce linear interpolation analysis between contrastive keywords as well as a similar triangular `linguistic' interpolation for an improved explainability of a text-to-image generation framework. 
\subsection{Discovering successful synthesized samples and \textit{Good} latent codes}
Given a fixed conditional text-to-image GAN model, the generator $G(z, (w, s))$ maps the latent space and the linguistic embeddings to the fake data distribution. It is well known that the synthetic sample space consists of high resolution pictures from ${Good}$ latent vectors as well as unreasonable images from ${Bad}$ latent codes. 
However, we only need successful generated samples and corresponding ${Good}$ latent codes for wide real-world applications.
In this subsection, we concentrate on proposing a framework for recognizing plausible images from numerous synthesized samples while deriving corresponding ${Good}$ latent vectors. 
\subsubsection{Pairwise linear interpolation of latent codes}
It has been extensively observed \cite{zhang2021divergan, shen2020interpreting} that when performing the linear interpolation between a successful starting-point latent vector and a successful end-point latent code, the appearance and the semantics of generated samples change continuously. In addition, DiverGAN \cite{zhang2021divergan} discovers that the generator is likely to synthesize a set of high-resolution pictures based on the pairwise linear interpolation between two ${Good}$ latent codes.  
This would imply that there may be close relation between successful synthesized images in the fake data space. That is to say, we may acquire a range of visually realistic pictures by sampling the latent vectors around a ${Good}$ latent code.

To further explore the semantic relationship between a plausible sample and an inadequate image in the synthetic image space, we visualize the samples generated by linearly interpolating a successful starting-point latent vector $z_{0}$ and an unsuccessful end-point latent code $z_{1}$.  
To be specific, the pairwise linear interpolation of latent codes is defined as:  
 \begin{equation}
 f(\gamma)=G((1-\gamma)z_{0}+\gamma z_{1}, (w, s)) \ \ for \ \ \gamma \in [0,1]
 \label{e1}
\end{equation}
where $\gamma$ is a scalar mixing parameter. In an attempt to quantitatively measure if there is a smooth transition from a perceptually plausible sample to an unsuccessful generated image, we calculate the learned perceptual image patch similarity (LPIPS) \cite{zhang2018unreasonable} score and the perceptual loss \cite{johnson2016perceptual} which reflect the diversity between two close interpolation samples.

We empirically observe that although the first and last part of interpolation results change gradually with the variations of the latent vectors, both the LPIPS score and the perceptual loss between intermediate samples are the largest and considerably increase, which we detail in Section \ref{5.2.1}. In other words, when linearly interpolating an unsuccessful latent code and a ${Good}$ latent vector,
the appearance and the semantics do not always vary smoothly along with the latent vectors. We therefore make the assumption that there exists a non-linear boundary separating successful generated images from unsuccessful synthesized samples in the fake image space. It implicitly means that image quality in the synthetic sample space may be distinguished. 
Suppose we have a non-linear image-quality function $f_{q} : \mathcal{X} \rightarrow t$, where $t$ represents the quality score. We are able to classify a synthesized sample as realistic or unsuccessful. 

\subsubsection{\textit{Good} \texorpdfstring{$\&$}{} \textit{Bad} data set creation}
\begin{figure*}
   \begin{minipage}[b]{1.0\linewidth}
   \centerline{\includegraphics[width=180mm]{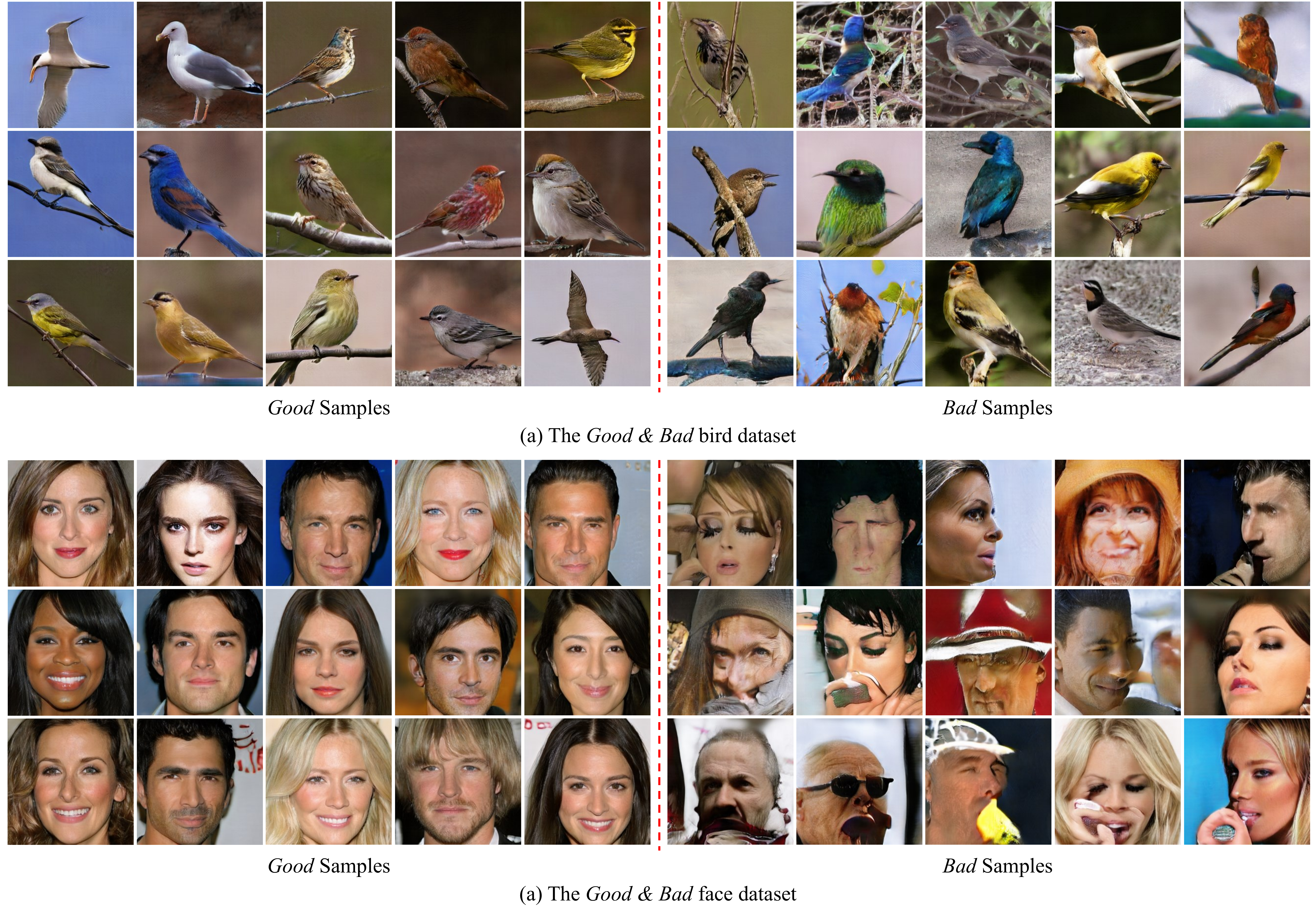}}
   \end{minipage}
   \caption{A snapshot of the {\em Good \& Bad} bird (\textbf{three top rows}) and face (\textbf{three bottom rows}) data sets: the \textbf{left} column is from the ${Good}$ data set; the \textbf{right} column is from the ${Bad}$ data set. These samples are synthesized by the recent DiverGAN generator \cite{zhang2021divergan}.}
   \vspace{-0.1in}
  \label{fig03} 
\end{figure*}

Our goal is to train a powerful classifier that can distinguish successful generated samples from unsuccessful synthetic images. To this end, we built two novel data sets (i.e., the {\em Good \& Bad} bird and face data sets) conditioned on the CUB bird data set \cite{wah2011caltech} and the Multi-Modal CelebA-HQ data set \cite{xia2021tedigan}, respectively. The {\em Good \& Bad} data set is a collection of perceptually realistic as well as implausible samples generated by a well-trained and fixed text-to-image GAN architecture.
The construction of the data set is based on a pilot study\footnote{In prep., 2022} on initial manual labeling (210+210 samples), which was used as the training set for automatic good vs bad binary classification. However, such a data set is too small to obtain a training set suitable for end-to-end, deep-learning based quality classification of generated images. Using strict criteria, an extended collection of mixed manual and automatic `good' and `bad' samples was constructed, within one day. 
Specifically, the used {\em Good \& Bad} bird data set consists of 6,700 synthesized samples, i.e., 2,700 ${Good}$ and 4,000 ${Bad}$ birds. The {\em Good \& Bad} face data set contains 2,000 successful generated faces as well as 2,000 unsuccessful synthetic faces.  
A summary of the {\em Good \& Bad} data set is reported in Table~\ref{tab:1}. 
 We visualize a snapshot of our data set in Fig. \ref{fig03}.
Below, we describe the procedure followed to construct the {\em Good \& Bad} data set.
\begin{table}
\caption{Statistics of the {\em Good \& Bad} bird and face data sets. `Bird' represents the {\em Good \& Bad} bird data set and `Face' denotes the {\em Good \& Bad} face data set.}
\begin{center}
\begin{tabular}{l c c c}
\hline
Dataset & train & test & total \\
\hline
Bird & 5,200& 1,500& 6,700\\
Face & 3,200& 800& 4,000\\
\hline
\end{tabular}
\end{center}
\label{tab:1}
\vspace{-0.15in}
\end{table}

\textbf{Image collection.} The first stage of creating the {\em Good \& Bad} data set involves producing a large set of candidate samples for each data set. The DiverGAN \cite{zhang2021divergan} has the ability to adopt a generator/discriminator pair to produce diverse, perceptually-plausible and semantically-consistent pictures, given a textual description and different injected noise on the latent vector. We therefore choose a pre-trained DiverGAN generator to acquire candidate images. We generated 30,000 synthesized samples as the basis for the selection of a {\em Good \& Bad} bird data set and a {\em Good \& Bad} face data set, respectively. 

\textbf{Image selection.}
Given a variety of candidate pictures, we choose images according to the following criteria:

1) A successful generated image is supposed to have vivid shape, rich color distributions, clear background as well as realistic details. For the face data set, photo-realistic images should also have pleasing, undistorted facial attributes (e.g., eyes, hair, makeup, head and mouth) and expressions. 

2) A synthetic picture with strange shape, blurry background or unclear color is viewed as ${Bad}$. Meanwhile, we reject faces with an implausible facial appearance or ornamentation (e.g., hat and glasses) as unsuccessful samples.

3) We exclude ambiguous images of the type where also for the human judge, the classification as ${Good}$ or ${Bad}$ is difficult. For instance, a bird with only a slightly strange body (e.g.,  lacking legs) is judged as an ambiguous-quality picture. 

\begin{figure*}
   \begin{minipage}[b]{1.0\linewidth}
   \centerline{\includegraphics[width=180mm]{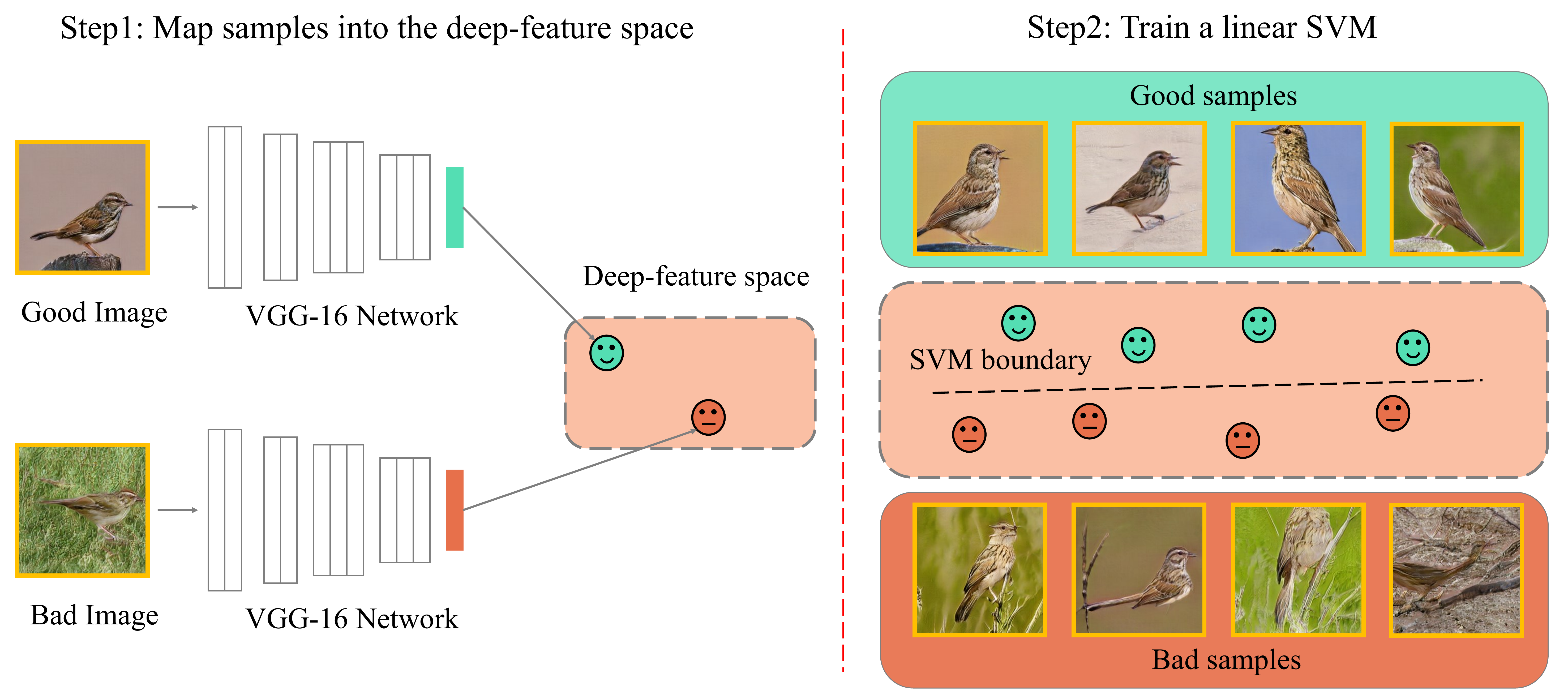}}
   \end{minipage}
   \caption{A schematic outline of the first two steps for automatically discovering ${Good}$ birds from the generated images.}
   \vspace{-0.1in}
  \label{fig04} 
\end{figure*}

For the {\em Good \& Bad} bird data set, we find it inefficient to manually choose thousands of plausible birds from 30,000 collected samples. To reduce the selection labor, we propose a process to obtain the desired birds as follows (depicted in Fig.~\ref{fig04}):  

1) Based on the principles mentioned before, we select 420 synthesized samples (i.e., 210 ${Good}$ and 210 ${Bad}$ birds) as the initial {\em Good \& Bad} bird data set, which is split into a training set (i.e., 150 ${Good}$ and 150 ${Bad}$ birds) and a testing set (i.e., 60 ${Good}$ and 60 ${Bad}$ birds). We intend to use these labeled samples to train a simple classification model to try to predict the quality class of synthesized images. However, it is difficult to directly apply a traditional classifier (e.g., a linear SVM) to separate realistic images adequately from inadequate samples, since the image instances exist in a non-linear manifold \cite{weinberger2006unsupervised}. In the meantime, we cannot train a deep neural network (e.g., VGG \cite{simonyan2014very}) from scratch to label a synthetic sample as ${Good}$ or ${Bad}$ due to the small number of the samples in the initial {\em Good \& Bad} training set. Bengio et al. \cite{bengio2013better} postulate that deep convolutional networks have the ability to linearize the manifold of pictures into a Euclidean subspace of deep features. Inspired by this hypothesis, we expect that ${Good}$ and ${Bad}$ samples can be classified by an approximately linear boundary in such deep-feature space.

2) We adopt the publicly available VGG-16 network trained on ImageNet to transform the image samples from the training set (i.e., 150 ${Good}$ and 150 ${Bad}$ samples) into the deep-feature representation of layer \verb+VGG-16/Conv5_1+. The obtained deep features and the corresponding labels (i.e., ${Good}$ and ${Bad}$) are used to fit a linear SVM model for automatic labeling of the samples in the deep-feature space. To evaluate the performance of the model, we transform the testing samples (i.e., 60 ${Good}$ and 60 ${Bad}$ birds) into deep-feature vectors while applying the learned SVM boundary to predict the classes for the unseen samples. 

3) In order to harvest an expanded set of Good or Bad samples, we use the trained SVM model to automatically label the 30,000 collected birds. We manually choose 2,700 ${Good}$ and 2,000 ${Bad}$ birds from the images that are classified as ${Good}$, which is not a laborious task due to the performance of the SVM. Moreover, to boost the diversity of ${Bad}$ birds on our data set, we select 2,000 ${Bad}$ birds from the samples that are predicted as ${Bad}$. Finally, 2,700 ${Good}$ and 4,000 ${Bad}$ birds are acquired as the final, expanded {\em Good \& Bad} bird data set. Also for the faces, we discovered that it is easy to label the synthesized samples as ${Good}$ or ${Bad}$. We therefore manually select 2,000 ${Good}$ and 2,000 ${Bad}$ samples from 30,000 synthetic faces for the {\em Good \& Bad} face data set. The manual selection was realized in one day.

\textbf{Splitting of the data set.}
The {\em Good \& Bad} face data set is randomly divided into the training and test sets with a ratio of 4:1. After the splitting, the training set comprises 3,200 images, i.e., 1,600 ${Good}$ and 1,600 ${Bad}$ faces. The test set consists of 800 samples including 400 ${Good}$ and 400 ${Bad}$ faces. The {\em Good \& Bad} bird data set contains 6,700 birds, where 5,200 images (i.e., 2200 ${Good}$ and 3,000 ${Bad}$ birds) belong to the training set and the other 1500 images (i.e., 500 ${Good}$ and 1,000 ${Bad}$ birds) belong to the test set. 
\subsubsection{Synthetic samples classification}
Given the extensive training set obtained in this manner, it is now possible to do the quality classification by end-to-end deep learning instead of using an unmodified, pretrained CNN and an SVM. To fully automatically distinguish successful synthesized samples from unrealistic images, we attempt to fine-tune a pre-trained CNN model (e.g., ResNet \cite{he2016deep}) on the proposed {\em Good \& Bad} data set, which we will detail in Section \ref{5.2.3}. We expect that this approach is able to achieve the best results. We therefore have the ability to effectively and efficiently identify photo-realistic samples from generated images while acquiring corresponding ${Good}$ latent vectors. These ${Good}$ latent codes can be exploited for further research, facilitating and extending the applicability of text-to-image generation in practice. For instance, we can produce a wealth of high-quality samples by conducting the pairwise linear interpolation between ${Good}$ latent codes, e.g., for the purpose of data augmentation. Given a ${Good}$ latent vector, we can synthesize several similar but semantically-diverse pleasing generated samples via latent-space navigation, which will be discussed in the next section. 

\subsection{Identifying meaningful latent-space directions}
In this subsection, we mathematically show that Semantic Factorization (SeFa) \cite{shen2021closed} approximately identifies the principal components, as PCA does. Furthermore, we propose a technique to capture semantically-interpretable latent-space directions for a conditional text-to-image GAN model. To optimize the edited sample, the background-flattening trick is presented to fine-tune the background.  
\subsubsection{Analyzing the correspondences between SeFa, GANSpace and PCA}
\label{4.2.1}
We attempt to discuss the relationship between SeFa \cite{shen2021closed} and GANSpace \cite{harkonen2020ganspace}, since they both introduce an algorithmically simple but surprisingly effective technique to derive semantically-understandable directions. Specifically,
\\GANSpace collects a set of latent codes and conducts PCA on them to identify the significant latent-space directions. SeFa proposes to directly decompose the pre-trained weights for semantic image editing. Mathematically, SeFa is formulated as:
\begin{equation}
A^{T}An_{i}-\lambda_{i}n_{i}=0
\label{e5}
\end{equation}
where $A\in R^{ d \times l}$ is the weight matrix of the first transformation step in the generator and $ \{{n_{i}}\}_{i=1}^{k}$ indicate $k$ most meaningful directions. The solutions to Equation \ref{e5} correspond to the eigenvectors of $ A^{T}A$ with respect to the $k$ largest eigenvalues. $A$ is usually normalized by L2 norm when implementing SeFa. The formulation of SeFa can almost be perceived as PCA on $A$, since the results of PCA are the eigen vectors of the covariance matrix $C_{A}$  associated with $A$ and $C_{A}$ is similar to $A^{T}A$. Specifically, $C_{A}$ is denoted as:
\begin{equation}
C_{A}=\frac{1}{d-1}{(A - <A>)}^{T}(A - <A>)
\end{equation}
where $<A>$ represents the mean from each column of $A$ and $C_{A}$ is the covariance matrix of $A$. 
The difference between regular PCA and SeFa is located in the normalization of $A$. We therefore argue that SeFa is approximately equivalent to regular PCA on the pre-trained weights. That is to say, GANSpace and SeFa perform PCA on the latent vectors and the pre-trained weights, respectively. 
\subsubsection{Independent component analysis for semantic discovery in the latent space}
It has been observed that the pre-trained weights of the standard GAN contain semantically-useful information. We can capture the meaningful latent-space directions in an unsupervised manner by exploiting the well-trained weights of the generator. A conditional text-to-image GAN generator typically leverages a dense layer to transform a latent code into a visual feature map, where a latent space is projected to another space and ultimately into an output image. We make the assumption that there exists a wealth of semantics in the initial fully-connected weight matrix of a text-to-image GAN model, due to the linguistic content of the text. We aim at presenting a simple algorithm extracting the main patterns of the pre-trained weights as the interpretable latent-space directions. More specifically, we hypothesize that when given the pre-trained weight matrix $A$ of the first linear layer of $G(z, (w, s))$, we can obtain a suite of $k$ meaningful semantic factors $N=(n_{1}, n_{2},\cdots, n_{k})$ by processing the weight matrix $A$. Mathematically,  
\begin{equation}
N=f(A)
\end{equation}
where $f(\cdot)$ is the function for semantic discovery. These acquired semantics should denote different attributes of the image. For example, $n_{1}$ represents pose, $n_{2}$ represents smile and $n_{3}$ represents gender for the face data set. To better manipulate the image generation, we argue that these components should be fully independent rather than just uncorrelated (orthogonal). However, when employing PCA as $f(\cdot)$ to discover the controllable latent-space directions, the obtained principal components are only uncorrelated, but not independent. 
Meanwhile, PCA is optimal for Gaussian data only \cite{ablin2018faster}, while the pre-trained weight matrix $A$ is not guaranteed to be Gaussian.
Here, we propose to utilize independent component analysis (ICA) to identify useful latent-space semantics for a conditional text-to-image GAN model.

The goal of ICA is to describe a $M\times L$ data matrix $X$ in terms of independent components. It is denoted as:
\begin{equation}
X=BS
\end{equation}
where $B$ is a $M\times T$ mixing matrix and $S$ is a $T\times L$ source matrix consisting of $T$ independent components. 

ICA is commonly viewed as a more powerful tool than PCA \cite{deng2012small}, since it is able to make use of higher-order statistical information incorporating a variety of significant features. Furthermore, ICA is adequate for analyzing non-Gaussian data.
To maximize both the independence and the orthogonality between the directions, i.e., $n_{1}, n_{2},\cdots, n_{k}$, we apply a fast ICA under an additional orthogonality constraint \cite{ablin2018faster} to directly decompose the pre-trained weight matrix to derive the meaningful directions in the latent space. The obtained vectors are therefore not only independent but also orthogonal. We expect that the components can lead to a more precise control over the latent space of the DiverGAN~\cite{zhang2021divergan} model.

\subsubsection{Background flattening}
A movement along an effective direction in the latent space should not only accurately change the desired attribute, but also maintain other image content, e.g., the background. However, when applying existing semantic-discovery methods even our introduced algorithm on the text-to-image generation model, we find that the background appearance in the edited sample usually varies along with the target attribute. To overcome this issue, we develop a {\em Background-Flattening Loss} (BFL) to fine-tune the acquired directions to improve the background.
 This loss is defined by using both low-level pixels and high-level features, ensuring that the background is optimized and other image contents are preserved. Specifically, it is denoted as:
\begin{equation}
\mathcal{L}_{flatten}(x_{1},x_{2})=||x_{1}-x_{2}||_{1} +\mathcal{L}_{LPIPS}(x_{1},x_{2})
\end{equation}
where $x_{1},x_{2}$ refer to a source sample and an edited sample, respectively. We leverage the Adam algorithm~\cite{Kingma2015AdamAM} to optimize the independent components. 

We empirically find that we are able to employ our proposed BFL to remove the patterns representing the background. To be specific, we can obtain a sample with a white background by increasing the distance (i.e., the BFL) between samples generated by different directions, since the white background and the black background will lead to the maximum loss values. After that, to remove the background, we take the white-background sample as the source image while reducing the distance between the source sample and the edited samples.
\subsection{Improving the explainability of the conditional text-to-image GAN}
\label{Norm}
In addition to the latent space, a conditional text-to-image GAN model also contains the linguistic embeddings, in which word and sentence vectors are adopted to module the visual feature map for semantic consistency. Despite high-quality pictures achieved by the existing approaches, we yet do not understand what a text-to-image generation architecture has learned within the linguistic space of the conditional input-text probes.

In order to understand `embeddings' in deep learning, several methods have been proposed. A common method is to visualize the space using, e.g., t-SNE or k-means clustering. This may give some insights on the location of dominant image categories in the sub space. An alternative approach is to utilize - yet another - step of dimensionality reduction by applying standard PCA on the embedding. However, this still does not lead to good explanations and an easy controllability of the image-generation process. In this subsection, we start from $Good$ latent vectors and introduce two basic techniques to provide insights into the explainability of a text-to-image synthesis framework.
\subsubsection{Linear interpolation and semantic interpretability}
We study the linear interpolation between a pair of keywords in order to qualitatively explore how well the generator exploits the linguistic space of the conditional input-text probes as well as testing the influence of individual, different words on the generated sample. We can observe how the samples vary as a word in the given text is replaced with another word, for instance by using a polarity axis of qualifier key words (dark-light, red-blue, ...). More specifically, we can first acquire two word embeddings (i.e., $w_{0}$ and $w_{1}$) and two corresponding sentence vectors (i.e., $s_{0}$ and $s_{1}$) by only altering a significant word (e.g., the color attribute value and the background value) in the input natural-language description. Afterwards, the results are obtained by performing the linear interpolation between the initial textual description $ (w_{0}, s_{0})$ and the changed description $ (w_{1}, s_{1})$ while keeping the $Good$ latent code $z$ frozen. Mathematically, this proposed text-space linear interpolation combines the latent code, the word and the sentence embeddings and is formulated as:
\begin{equation}
h(\gamma)=G(z, (1-\gamma)w_{0}+\gamma w_{1}, (1-\gamma)s_{0}+\gamma s_{1}) 
\end{equation}
where $\gamma \in [0,1]$ is a scalar mixing parameter and $z$ is a successful latent code.

For the CUB bird dataset, when we vary the color attribute value in the given sentence, we empirically explore what happens in the color mix: Do we, e.g., get an average color interpolation in RGB space or does the network find another solution for the intermediate points between two disparate embeddings? 

In general, our presented text-space linear interpolation has the following advantages: 

\noindent
$\bullet$~The linear interpolation between a pair of keywords can be utilized to quantitatively control the attribute of the synthetic sample, when the attribute varies smoothly with the variations of the word vectors. For example, the length of the beak of a bird can be adjusted precisely via the text-space linear interpolation between the word embeddings of `short' and `long'. 

\noindent
$\bullet$~When the attribute of the synthesized sample does not change gradually along with the word embeddings, we can exploit a text-space linear interpolation to produce a variety of novel samples. Take bird synthesis as an example: When conducting the linear interpolation between color keywords, 
\\$G(z, (w, s))$ is likely to generate a new bird whose body contains two colors (e.g., red patches and blue patches) in the middle of the interpolation results, as shown in Fig.~\ref{fig12}. 

\noindent 
$\bullet$~Through the linear interpolation between contrastive keywords, we can take a deep look into which keywords play important roles in yielding foreground images as well as which image (background) regions are determined by the terms in the text probe.
\subsubsection{Triangular interpolation and semantic interpretability}
We extend the pairwise linear interpolation between two points to the interpolation between three points, i.e., in the 2-simplex, for further studying $G(z, (w, s))$ and better performing data augmentation. Since this kind of interpolation forms a triangular plane, we name it the triangular interpolation. The triangular interpolation is able to generate more and more diverse samples conditioned on three corners (e.g., latent vectors and keywords), spanning a field rather than a line.

Similar to the linear interpolation between a pair of keywords, we need to derive three word embeddings (i.e., $w_{0}$, $w_{1}$ and $w_{2}$) and three corresponding sentence vectors (i.e., $s_{0}$, $s_{1}$ and $s_{2}$) as corners to define the presented text-space triangular interpolation:
\begin{equation}
\begin{split}
h(\gamma_{1}, \gamma_{2})=G(z, &(1-\gamma_{1}-\gamma_{2})w_{0}+\gamma_{1} w_{1}+\gamma_{2} w_{2}, \\
&(1-\gamma_{1}-\gamma_{2})s_{0}+\gamma_{1} s_{1}+\gamma_{2} s_{2})
\end{split}
\end{equation}
where $\gamma_{1} \in [0,1]$ and $\gamma_{2} \in [0,1]$ are mixing scalar parameters and $z$ is a successful latent vector. 

For the sake of attribute analysis, we can obtain three new textual descriptions by replacing the attribute word in the initial natural-language description with another two attribute words. Then, through the triangular interpolation between keywords, the generator has the ability to yield pictures based on the above three attributes. Moreover, we expect that the text-space triangular interpolation should achieve the same visual smoothness as the text-space linear interpolation. In other words, when fixing the weight (i.e., $\gamma_{2}$) of the third text in  the triangular interpolation between keywords, the attributes of the image vary gradually along with the word embeddings if the interpolation results of a text-space linear interpolation between the first two textual descriptions change continuously.

The text-space triangular interpolation has obvious advantages over the linear interpolation between a pair of keywords. Firstly, the text-space triangular interpolation is able to produce more image variation to perform data augmentation than the pairwise linear interpolation. Secondly, we can simultaneously control two different attributes (e.g., color and the length of the beak) via  the triangular interpolation between keywords. Thirdly, through the text-space triangular interpolation, three identical attributes (e.g., red, yellow and blue) can be combined to synthesize a novel sample.
\section{Experiments}
\label{er}
\subsection{Experimental settings}
\label{es}
\textbf{Datasets.} We perform a set of experiments on three broadly utilized text-to-image data sets, i.e., the CUB bird \cite{wah2011caltech}, MS COCO \cite{lin2014microsoft} and Multi-Modal
CelebA-HQ \cite{xia2021tedigan} data sets. 

$\bullet$ \textbf{CUB bird.} The CUB bird data set contains a total of 11,788 images, in which 8,855 images are taken as the training set and the remaining 2,933 images are employed for testing. Each bird is associated with 10 textual descriptions.
\begin{figure*}
   \begin{minipage}[b]{1.0\linewidth}
   \centerline{\includegraphics[width=180mm]{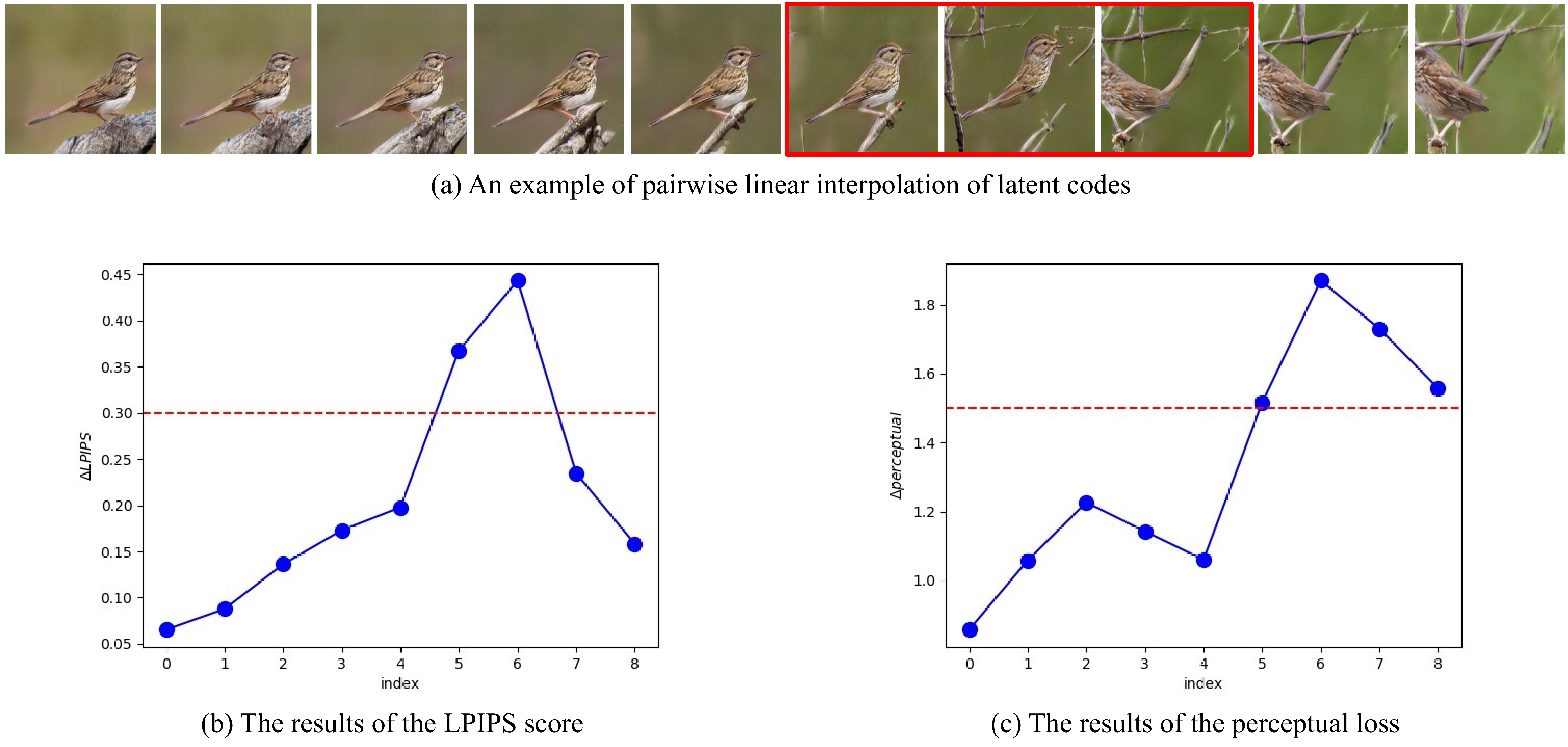}}
   \end{minipage}
   \caption{An example of the pairwise linear interpolation of latent vectors (${Good}$ $\rightarrow$ ${Bad}$). The red bounding box in (a) emphasizes a discontinuous range within the linear-interpolation results. The dashed red line in (b) and (c) is an approximate boundary distinguishing smooth changes from discontinuous variations, determined by our observations. The index number represents the comparison, starting with 0, i.e., the comparison between the first and the second image on the left. The discontinuity is quantitatively revealed both in LPIPS and in perceptual loss.}
   \vspace{-0.1in}
  \label{fig05} 
\end{figure*}

$\bullet$ \textbf{MS COCO.} The MS COCO data set is a more challenging data set consisting of 123,287 images in total, which are split into 82,783 training pictures and 40,504 test pictures. Each image includes 5 human annotated captions. 

$\bullet$ \textbf{Multi-Modal CelebA-HQ.} The Multi-Modal CelebA-HQ data set is composed of 24,000 and 6,000 faces for training and testing, respectively. Each face is annotated with 10 sentences.

\textbf{Implementation details.} We take the recent DiverGAN generator \cite{zhang2021divergan} as the backbone generator, which is pre-trained on the CUB bird, Multi-Modal CelebA-HQ and MS COCO data sets. 
The image size of the proposed {\em Good \& Bad} data set is set to $256 \times 256 \times 3$. We set the output dimension of the CNN models (e.g., ResNet \cite{he2016deep} and VGG \cite{simonyan2014very}) to 2. We adopt the Adam optimizer \cite{Kingma2015AdamAM} with a batch size of 64 to fine-tune the classification network pre-trained on ImageNet. We utilize the learning-rate finder technique \cite{smith2017cyclical} to acquire a suitable learning rate. The one cycle learning rate scheduler \cite{smith2019super} is leverage to dynamically alter the learning rate whilst the model is training. 
We set the manipulation intensity $\alpha$ to 3 for SeFa \cite{shen2021closed} and our proposed algorithm. The scalar parameter for GANSpace \cite{harkonen2020ganspace} is set to 20 on the CUB bird data set and 9 on the COCO data set, respectively. We employ the Adam optimizer with $\beta=(0.0, 0.9)$ to fine-tune the identified directions. 
We set the learning rate to 0.0001.
The steps of a linear interpolation are set to 10. We set the steps of $\gamma_{1}$ and $\gamma_{2}$ in a triangular `linguistic' interpolation to 10. 
Our methods are implemented by PyTorch~\cite{paszke2019pytorch}.
We conduct all the experiments on a single NVIDIA Tesla V100 GPU (32 GB memory).
\subsection{Results of finding \textit{Good} synthetic samples}
\label{cws}
\subsubsection{Results of the pairwise linear interpolation of latent codes}
\label{5.2.1}
To better understand the transition process from a successful synthesized sample to an unsuccessful generated image, we visualize the results of the pairwise linear interpolation between a ${Good}$ latent code and a ${Bad}$ latent vector in Fig.~\ref{fig05} (a). It can be observed that for the first five and the last two pictures, both the background and the visual appearance of footholds vary gradually along with the latent vectors. However, the background, the visual appearance of footholds, the positions, the shapes and even the orientations ($7^{th}\rightarrow8^{th}$ sample) of the birds do not change continuously from the $6^{th}$ image to the $8^{th}$ sample. It suggests that there may exist a non-linear boundary separating ${Good}$ samples from ${Bad}$ images in the fake data space.    

We also show the corresponding LPIPS score and the perceptual loss (presented in Fig.~\ref{fig05} (b) and Fig.~\ref{fig05} (c)) to quantitatively compare the diversity between two close samples. It can be seen that the increase of the $6^{th}$ point ($6^{th}\rightarrow7^{th}$ sample) is the largest and the $7^{th}$ point ($7^{th}\rightarrow8^{th}$ sample) obtains the highest score for both the LPIPS and the perceptual loss. Meanwhile, both points are over the red line which is an approximate boundary distinguishing smooth changes from discontinuous variations and determined by our observations. The results of Fig.~\ref{fig05} (b) and Fig.~\ref{fig05} (c) match what observe in Fig.~\ref{fig05} (a), indicating that the visual appearance of the birds does not always vary smoothly along with the latent codes.
\begin{table}
\begin{center}
\begin{tabular}{c c}
\hline
Methods & Accuracy($\%$) \\
\hline
Image& 70.0 \\
PCA-Image & 73.3 \\
Latent Code & 75.8 \\
VGG-16(conv5$\_$3)& 94.2\\
VGG-16(conv5$\_$2)& 96.7\\
\hline
$\textbf{VGG-16(conv5$\_$1)}$& $\textbf{97.5}$\\
\hline
\end{tabular}
\end{center}
\caption{Classification accuracy on the separation boundary with respect to image quality. {\em Image} refers to a direct application of SVM on the image pixels. {\em PCA-Image} refers to using PCA on the image pixels after reducing the dimensionality to 128 and applying SVM to identify realistic samples. {\em Latent Code} refers to the direct application of SVM in the latent space.}
\label{tab:2}
\end{table}
\begin{figure}[t]
  \begin{minipage}[b]{1.0\linewidth}
  \centerline{\includegraphics[width=85mm]{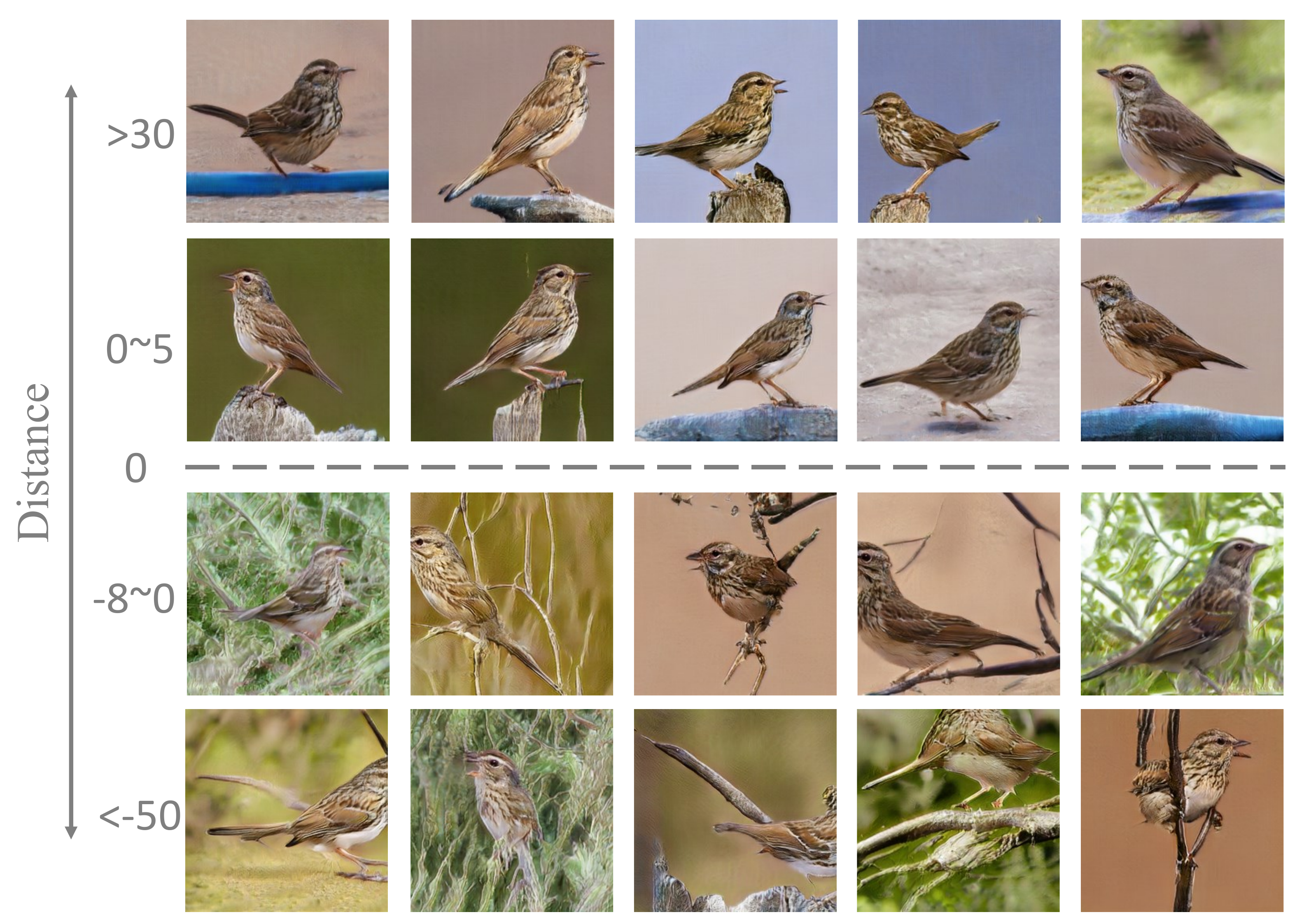}}
  \end{minipage}
  \caption{Example of partitioning of latent-code space between ${Good}$ (two top rows) and ${Bad}$ latent codes (two bottom rows), as determined by the discriminant value (distance) computed by a linear SVM (training set: Ngood=150, Nbad=150.)}
  \vspace{-0.1in}
  \label{fig06} 
\end{figure}
\begin{table}
\caption{Classification performance of the deep convolutional networks on the {\em Good \& Bad} bird and face data sets. ${Bird}$ refers to the {\em Good \& Bad} bird data set and ${Face}$ refers to the {\em Good \& Bad} face data set. The best results are in bold.} 
\begin{center}
\begin{tabular}{l c c}
\hline
\multicolumn{3}{c}{Classification performance/$\%$ }\\
\hline
\multirow{2}*{Method} & \multicolumn{2}{c}{Dataset} \\ 
\cline{2-3}
& Bird  & Face \\  
\hline
VGG-11 & 96.53& 98.08\\
VGG-16 & 95.70& 97.84\\
VGG-19 & 97.85& 98.20\\
ResNet-18 & 97.59& 98.68\\
ResNet-50 & \textbf{98.09}& 98.56\\
ResNet-101 & 97.79& \textbf{99.16}\\ 
\hline
\end{tabular}
\end{center}
\label{tab:3}
\vspace{-0.15in}
\end{table}
\begin{figure*}
   \begin{minipage}[b]{1.0\linewidth}
   \centerline{\includegraphics[width=180mm]{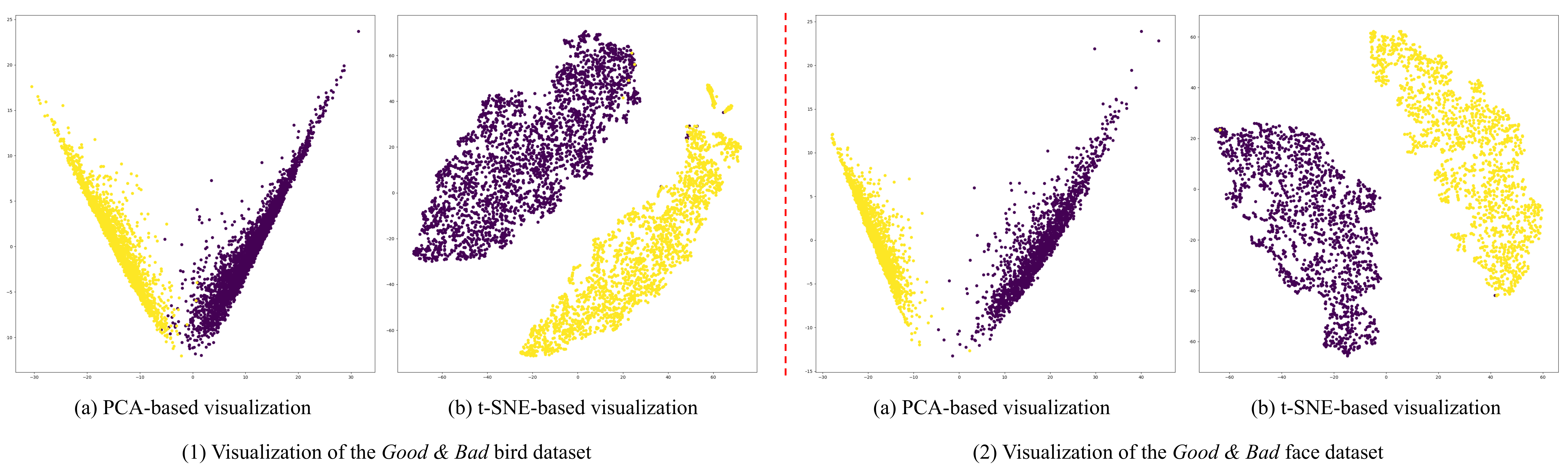}}
   \end{minipage}
   \caption{The visualization for the samples on the {\em Good \& Bad} data set by utilizing the PCA \cite{wold1987principal} and t-SNE \cite{van2008visualizing} methods. In this figure, the yellow color represents the ${Good}$ sample and the purple color represents the ${Bad}$ image.}
   \vspace{-0.1in}
  \label{fig07} 
\end{figure*}
\begin{figure*}
   \begin{minipage}[b]{1.0\linewidth}
   \centerline{\includegraphics[width=165mm]{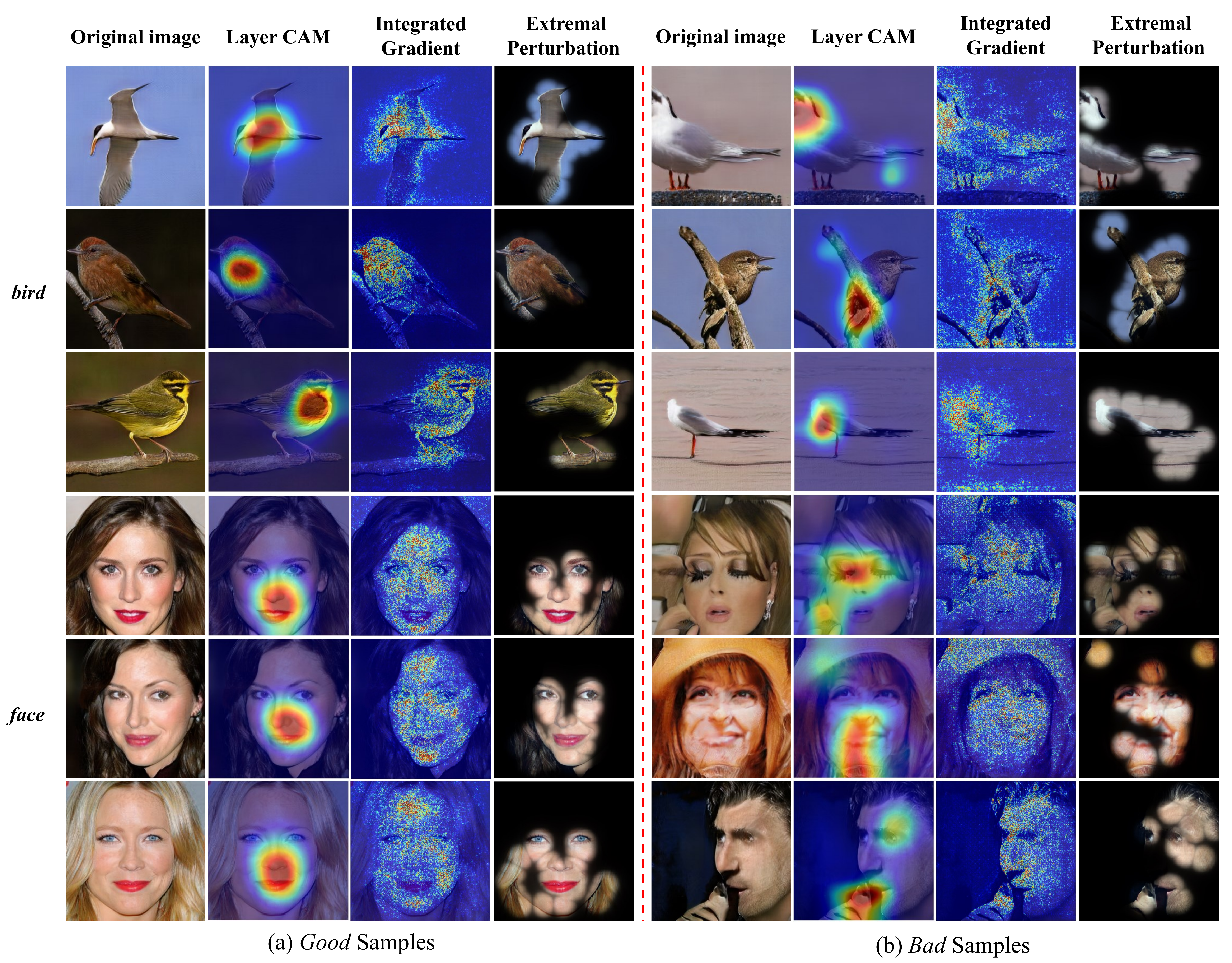}}
   \end{minipage}
   \caption{Explaining the image classification prediction made by ResNet-50 on the {\em Good \& Bad} bird data set (three top rows) and ResNet-101 on the {\em Good \& Bad} face data set (three bottom rows) using Layer-CAM \cite{jiang2021layercam}, integrated gradient \cite{sundararajan2017axiomatic} and extremal perturbation \cite{fong2019understanding}. The \textbf{left} half of the grid is from the ${Good}$ data set; the \textbf{right} half of the grid is from the ${Bad}$ data set, separated by the dashed line.}
   \vspace{-0.1in}
  \label{fig08} 
\end{figure*}
\subsubsection{Results on the initial \textit{Good} \texorpdfstring{$\&$}{} \textit{Bad} bird data set}
We try different methods to classify a synthetic sample as ${Good}$ or ${Bad}$ on the initial {\em Good \& Bad} bird data set (i.e., 210 ${Good}$ and 210 ${Bad}$ birds). The results are reported in Table~\ref{tab:2}. Here, we discover that all methods using the learned feature vectors of a well-trained VGG-16 network achieve over 94$\%$, suggesting that there exists a (almost) linear boundary in the deep-feature space which can accurately distinguish ${Good}$ samples from ${Bad}$ samples. In addition, the conv5\_1 activation in the pre-trained network obtains the best performance (accuracy: 97.5$\%$). We also attempted to employ the SVM with radial basis function (RBF) kernel to classify deep features, acquiring the same result as the linear SVM. Moreover, it can be observed that directly operating on the image pixels (accuracy: 70.0$\%$) and the latent space (accuracy: 75.8$\%$) does not work well for the classification of ${Good}$ and ${Bad}$ samples/latent codes. To boost the accuracy, we conduct PCA on the image pixels to reduce the dimension to 128 and apply a linear SVM to identify realistic samples. However, the accuracy is only improved by 3.3$\%$. The above results confirm the effectiveness of our proposed framework. 

We visualize some typical output samples selected from the test set (Ngood=60, Nbad=60) in Fig.~\ref{fig06} according to their distance to the decision boundary of the trained SVM. It can be observed that ${Good}$ samples are distinguishable from ${Bad}$ samples. Meanwhile, the ${Bad}$ birds around the boundary may have higher quality than the ${Bad}$ birds far from the decision boundary. It should be noted that in non-ergodic problems, where there is not a natural single signal source for the ${Good}$ (or the ${Bad}$) images, but there rather exists a partitioning of space, the SVM discriminant value for a sample is not guaranteed to be consistent with the intuitive prototypicality of the heterogeneous underlying class~\cite{van2014separability} due to the lack of a central density for that class.
\subsubsection{Results on the \textit{Good} \texorpdfstring{$\&$}{} \textit{Bad} data set}
\label{5.2.3}
\noindent\textbf{The classification results.} We fine-tune the pre-trained CNN models (i.e., ResNet and VGG) on the {\em Good \& Bad} data set in order to accurately predict the quality classes of generated images. The comparison between VGG-11, VGG-16, VGG-19, ResNet-18, ResNet-50 and Res-Net-101 with respect to the classification performance on the {\em Good \& Bad} bird and face data sets is shown in Table~\ref{tab:3}. We can observe that ResNet-50 achieves the best result (accuracy: 98.09$\%$) on the {\em Good \& Bad} bird data set and  ResNet-101 impressively acquires the accuracy of 99.16$\%$ on the {\em Good \& Bad} face data set. It can also be seen that ResNet performs better that VGG  and all the networks obtain a better than 95$\%$ accuracy on both the {\em Good \& Bad} bird data set and the {\em Good \& Bad} face data set. The above results demonstrate that the ${Good}$ and ${Bad}$ samples in the synthetic image space can be effectively distinguished by a well-trained deep convolutional network. 

\noindent\textbf{Visualization of the learned representation.} To visually investigate the distribution of the features learned by the CNN models (i.e., ResNet-50 for the {\em Good \& Bad} bird data set and ResNet-101 for the {\em Good \& Bad} face data set), we exploit the PCA \cite{wold1987principal} and t-SNE \cite{van2008visualizing} approaches to embed the samples on the {\em Good \& Bad} data set into a 2-dimensional space as shown in Fig.~\ref{fig07}. From this figure, we can see that the learned representations of the classification networks from different classes (i.e., ${Good}$ and ${Bad}$) are well separated indicating that the image classification models can project the plausible and unrealistic samples into two diverse latent spaces. Therefore, discovering photo-realistic samples from synthesized images is feasible. It can also be observed that 
the samples of different categories on the {\em Good \& Bad} face data set are more scattered than the {\em Good \& Bad} bird data set, which demonstrates that ResNet-101 trained on the {\em Good \& Bad} face data set performs better than ResNet-50 trained on the {\em Good \& Bad} bird data set. In other words, faces are easier to recognize than birds, which is consistent with the classification score.    

\noindent\textbf{Explaining the classification prediction.} We leverage three different methods (i.e., Layer-CAM \cite{jiang2021layercam}, integrated gradient \cite{sundararajan2017axiomatic} and extremal perturbation \cite{fong2019understanding}) to explain the image classification prediction obtained by ResNet-50 trained on the {\em Good \& Bad} bird data set and ResNet-101 trained on the {\em Good \& Bad} face data set.
Fig.~\ref{fig08} shows the explanation for the top 1 predicted class, suggesting that the classification network derives the results by concentrates on the discriminative regions of the objects (i.e., birds and face). For instance, Layer-CAM visualization ($2^{nd}$ and $6^{th}$ column) localizes the heads and belly of the birds and the noses, mouths and eyes of the faces. Meanwhile, integrated gradient  ($3^{rd}$ and $7^{th}$ column) and extremal perturbation ($4^{th}$ and $8^{th}$ column) correctly highlight the branches and the whole bodies of the birds while capturing the hat and the entire faces, pinpointing the reason why the samples are classified into the corresponding categories. More importantly, the blurry regions of the images ($6^{th}$, $7^{th}$ and $8^{th}$ column) are accurately identified by these explainable approaches. That is to say, our classification model can separate implausible regions from high-quality patches and discover successful synthetic samples from generated images.

\subsection{Results of latent-space manipulation}
\label{5.3}
\subsubsection{Comparison between SeFa, GANSpace and PCA}
\begin{figure}[t]
  \begin{minipage}[b]{1.0\linewidth}
  \centerline{\includegraphics[width=85mm]{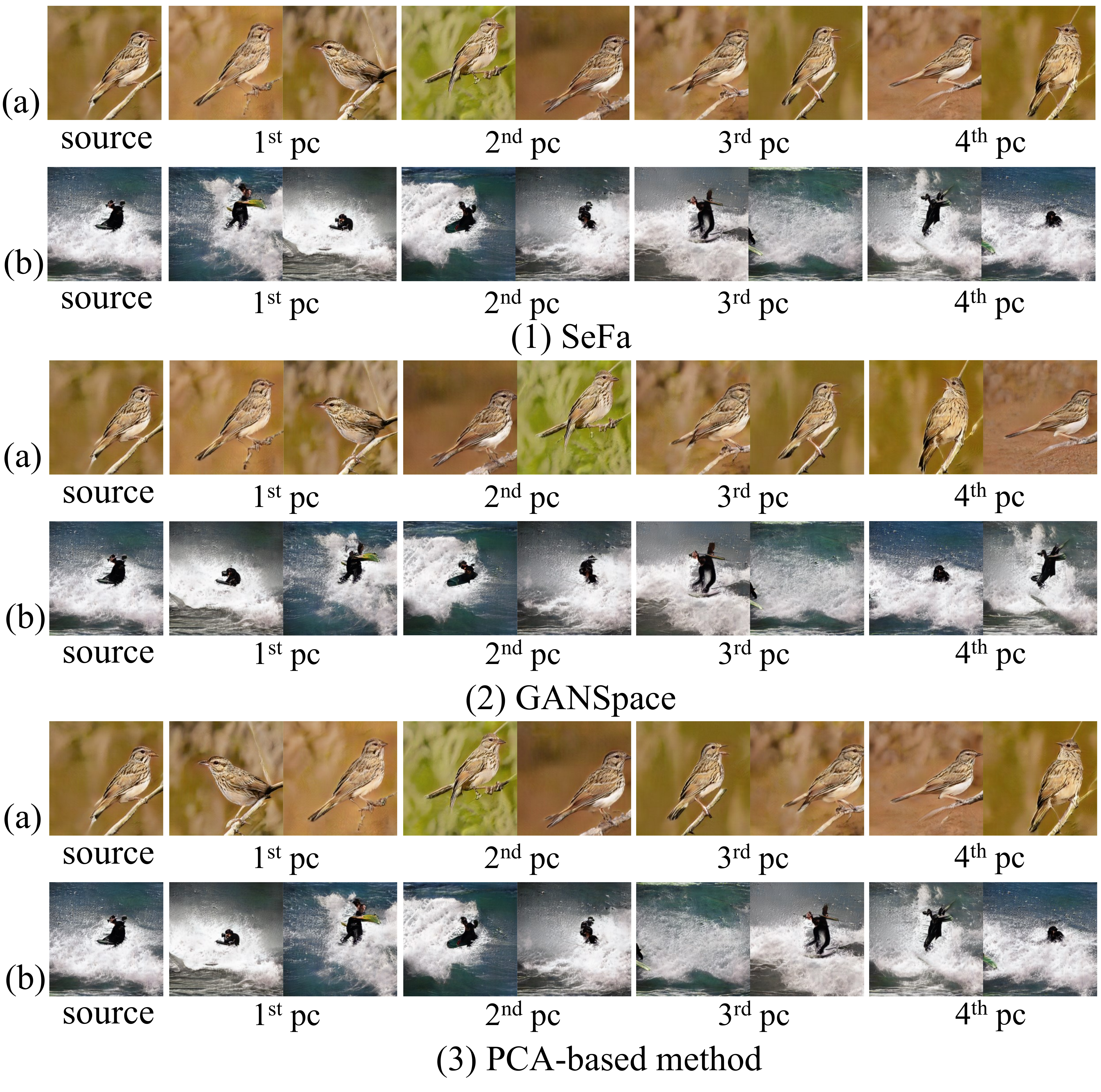}}
  \end{minipage}
  \caption{Visualization of individual components within the latent codes, for (1) SeFa \cite{shen2021closed}, (2) GANSpace \cite{harkonen2020ganspace} and (3) regular PCA. The original source image is in the left column (2 examples, a and b). For each principal component (pc1-pc4), example images from the negative and the positive side of its axis are shown.}
  \vspace{-0.1in}
  \label{fig09} 
\end{figure}

Fig.~\ref{fig09} plots the latent-code manipulation results of SeFa \cite{shen2021closed}, GANSpace \cite{harkonen2020ganspace} and regular PCA on the CUB bird and COCO data sets. We discover that these three approaches derive almost the identical directions although for some components (e.g., $4^{th}$ principal component) the negative and the positive side is reversed, supporting our claim in Section \ref{4.2.1}. Note that GANSpace is implemented by leveraging the first dense layer of DiverGAN to collect 10,000 sets of feature maps while performing PCA on them to obtain principal components as useful attributes. Additionally, we adjust the max manipulation intensity (i.e., $\alpha$ in Section \ref{preliminary}) to 20 on the CUB bird data set and 9 on the COCO data set, respectively. The above analysis suggests that when enough data is sampled, SeFa is similar to GANSpace for DiverGAN.
\begin{figure*}
   \begin{minipage}[b]{1.0\linewidth}
   \centerline{\includegraphics[width=160mm]{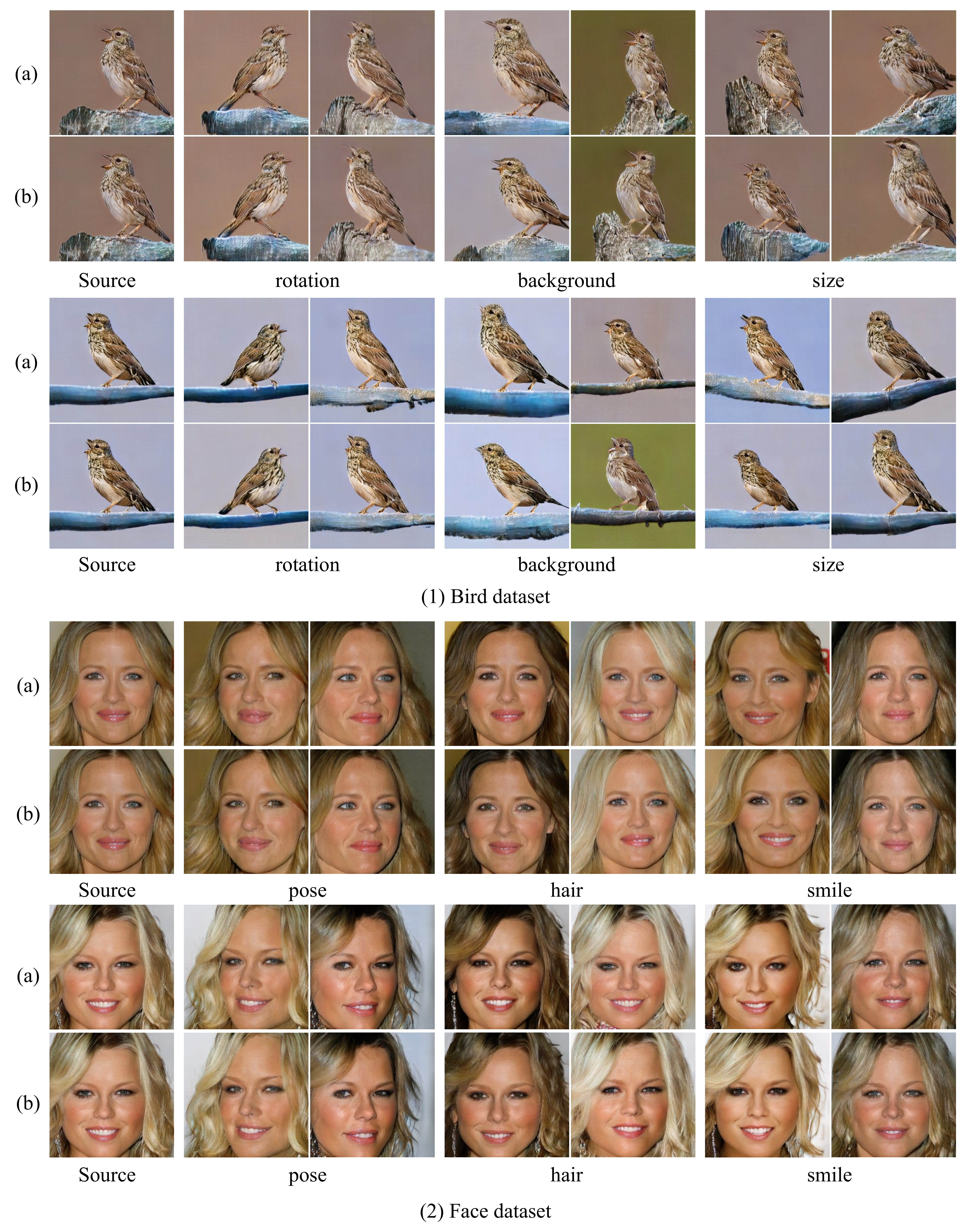}}
   \end{minipage}
   \caption{Qualitative comparison of the meaningful latent-space directions discovered by (a) SeFa \cite{shen2021closed} and (b) our proposed algorithm on (1) the CUB bird (four top rows) and (2) Multi-Modal CelebA-HQ (four bottom rows) data sets.}
   \vspace{-0.1in}
  \label{fig10} 
\end{figure*}
\subsubsection{Comparison with unsupervised methods}
For qualitative comparison, we visualize the meaningful directions identified by our proposed algorithm and SeFa on the CUB bird and Multi-Modal CelebA-HQ data sets in Fig.~\ref{fig10}. We can tell that our method is able to derive several fine-grained semantics corresponding to rotation, background and size for the bird model and pose, hair and smile for the face model, validating its effectiveness. Meanwhile, our approach leads to a more powerful control over the latent codes than SeFa. For example, when editing the background on the CUB bird data set and the smile on the Multi-Modal CelebA-HQ data set, our algorithm better preserves the size of the bird and the pose of the face, respectively.  It can also be seen that our method captures the same rotation and pose attributes as SeFa. The reason for this may be that ICA under orthogonal constraint and PCA can discover exactly the same most representative semantics (rotation for the bird model and pose for the face model).   The above results demonstrate that based on the ${Good}$ latent codes found by our well-trained classification model, we can adopt our presented algorithm to acquire a wealth of semantically-diverse and perceptually-realistic samples.
\begin{figure}[t]
  \begin{minipage}[b]{1.0\linewidth}
  \centerline{\includegraphics[width=85mm]{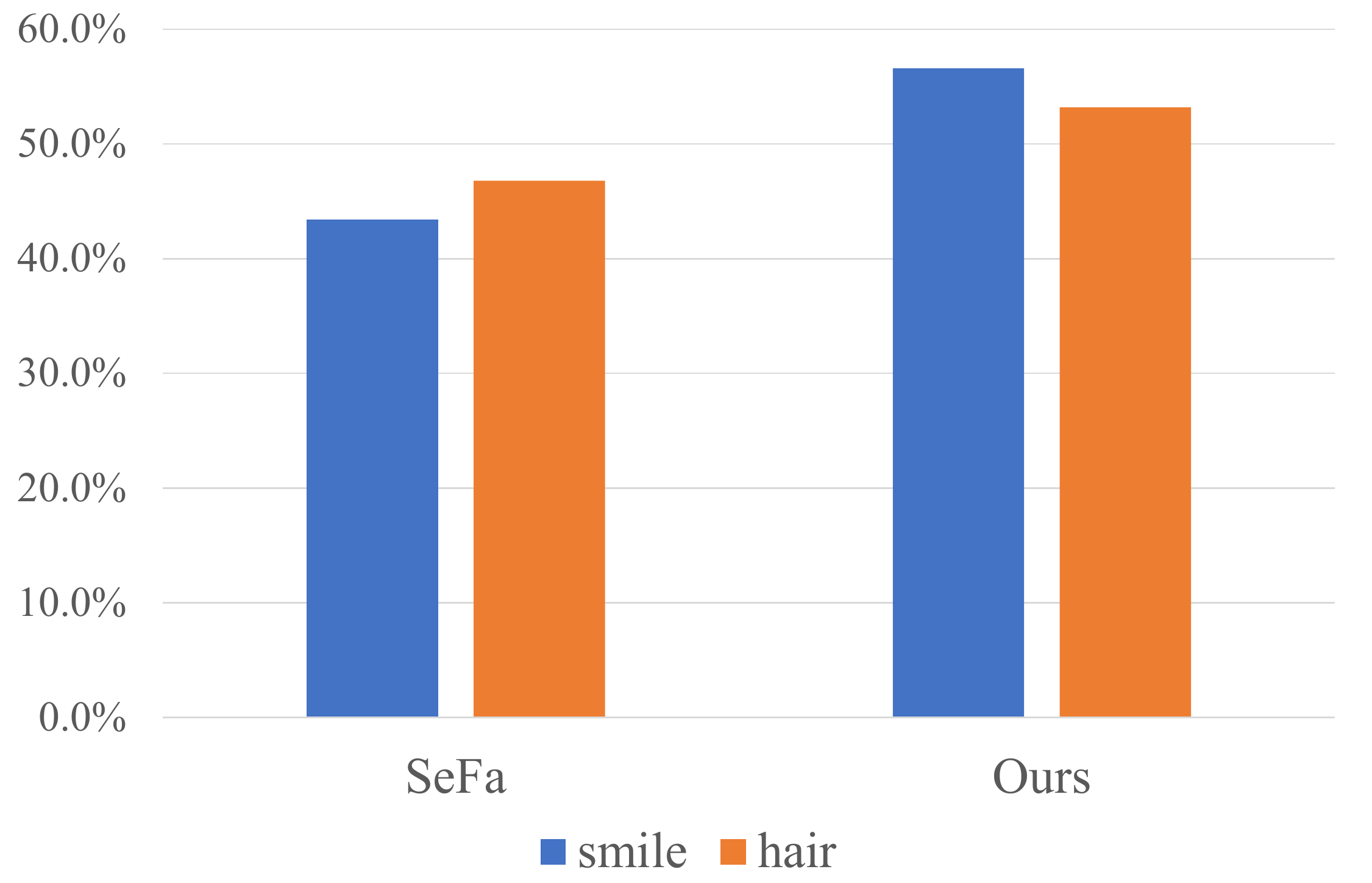}}
  \end{minipage}
  \caption{Human test results (ratio of 1st) of SeFa  \cite{shen2021closed} and our proposed method with respect to the smile and hair semantics on the Multi-Modal CelebA-HQ data set.}
  \vspace{-0.1in}
  \label{ht} 
\end{figure}
\begin{figure}[t]
  \begin{minipage}[b]{1.0\linewidth}
  \centerline{\includegraphics[width=85mm]{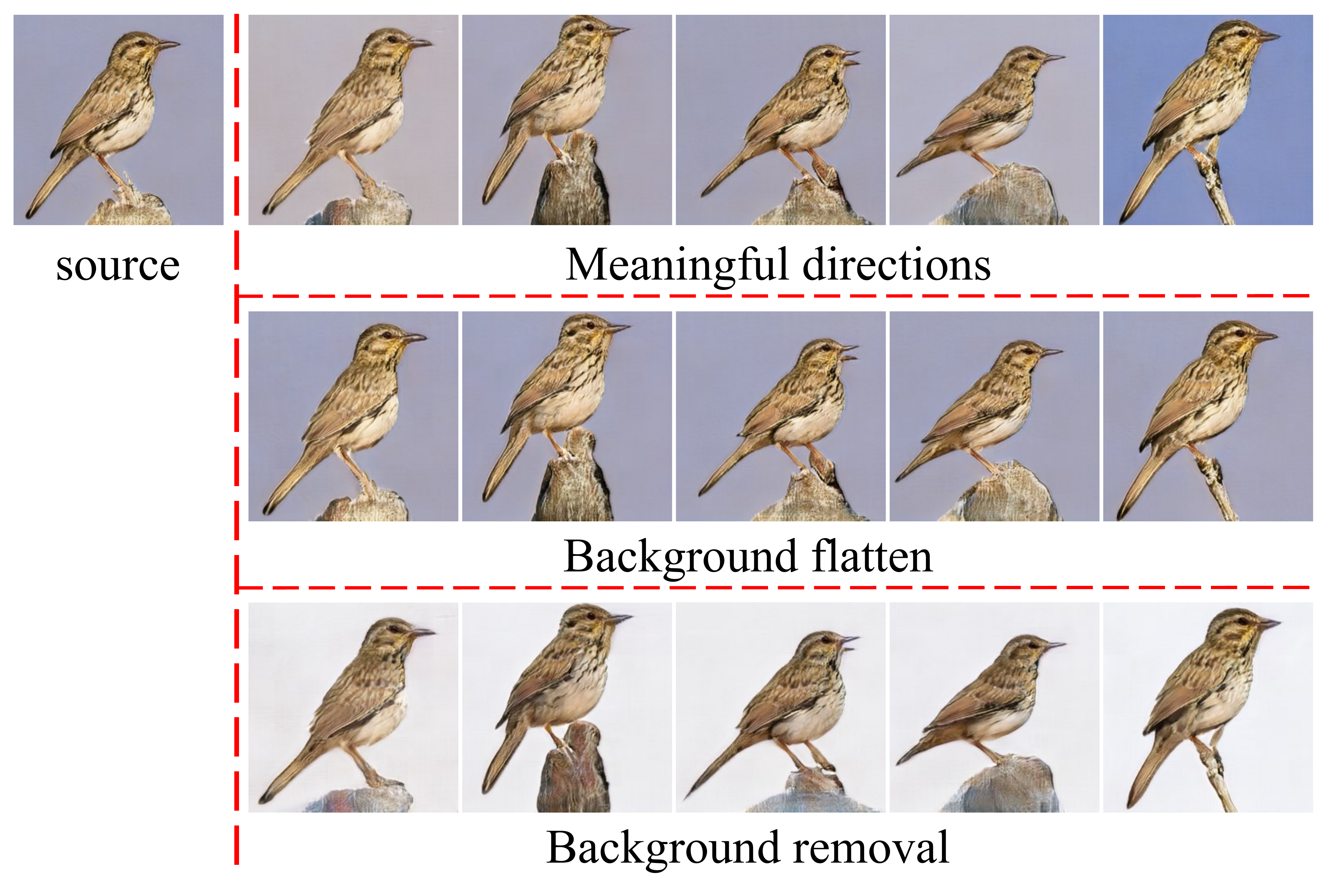}}
  \end{minipage}
  \caption{Visualization of background flatten and background removal for the meaningful directions acquired by our proposed method.}
  \vspace{-0.2in}
  \label{fig11} 
\end{figure}
\subsubsection{Human evaluation}
We conduct a human test on the Multi-Modal CelebA-HQ data set to compare our method with SeFa. We randomly select 100 successful synthesized faces while employing the directions (i.e., smile and hair) found by these two approaches to edit them. Users are asked to choose the sample with the most accurate change. Simultaneously, the final results are calculated by two judges for fairness. 
As illustrated in Fig.~\ref{ht}, our method performs better than SeFa with respect to the control of smile and hair, which demonstrates the superiority of our proposed algorithm. 
\begin{figure}[t]
  \begin{minipage}[b]{1.0\linewidth}
  \centerline{\includegraphics[width=85mm]{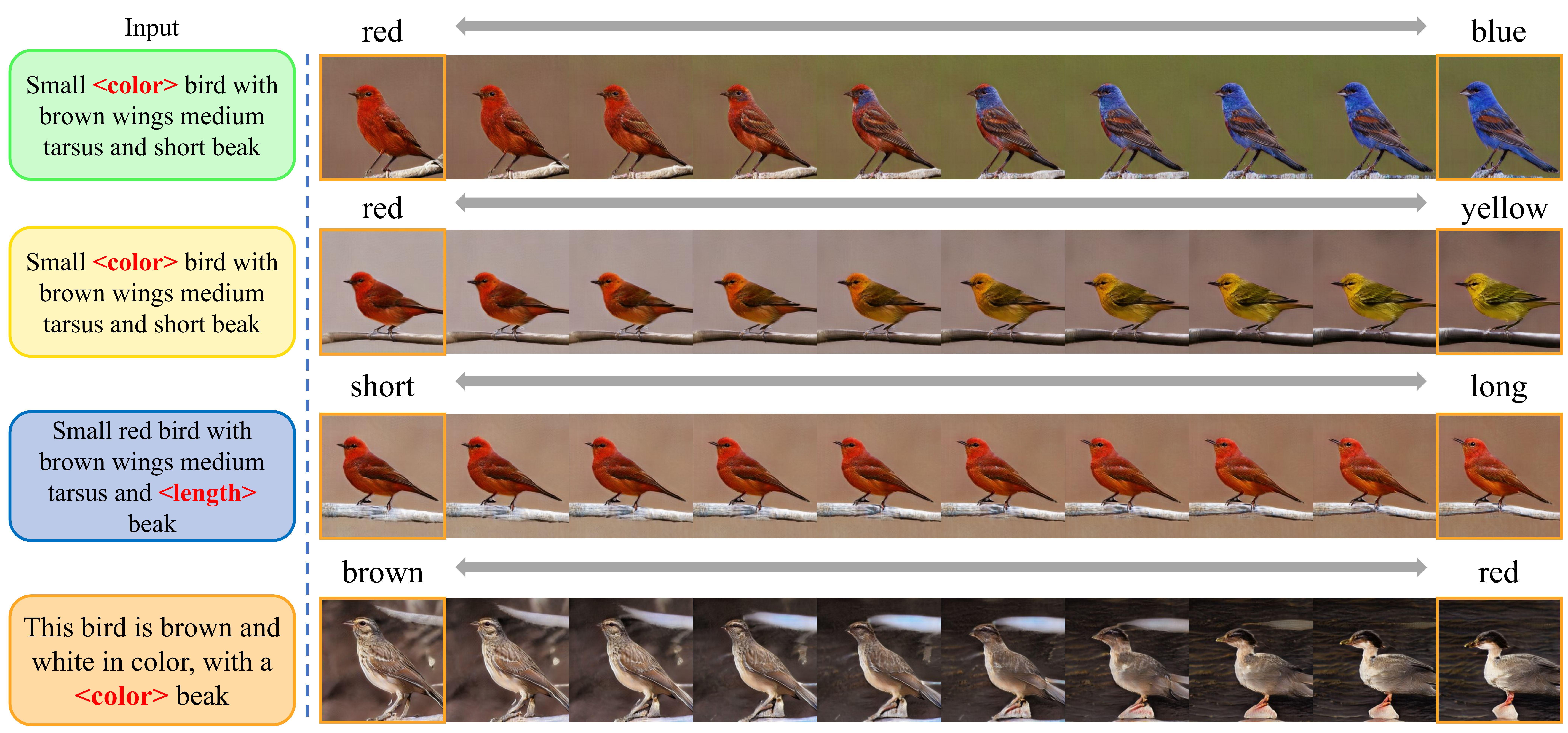}}
  \end{minipage}
  \caption{`Linguistic' interpolation of DiverGAN random latent-code samples on the CUB dataset, for four text input probes.}
  \vspace{-0.1in}
  \label{fig12} 
\end{figure}
\begin{figure}[t]
  \begin{minipage}[b]{1.0\linewidth}
  \centerline{\includegraphics[width=85mm]{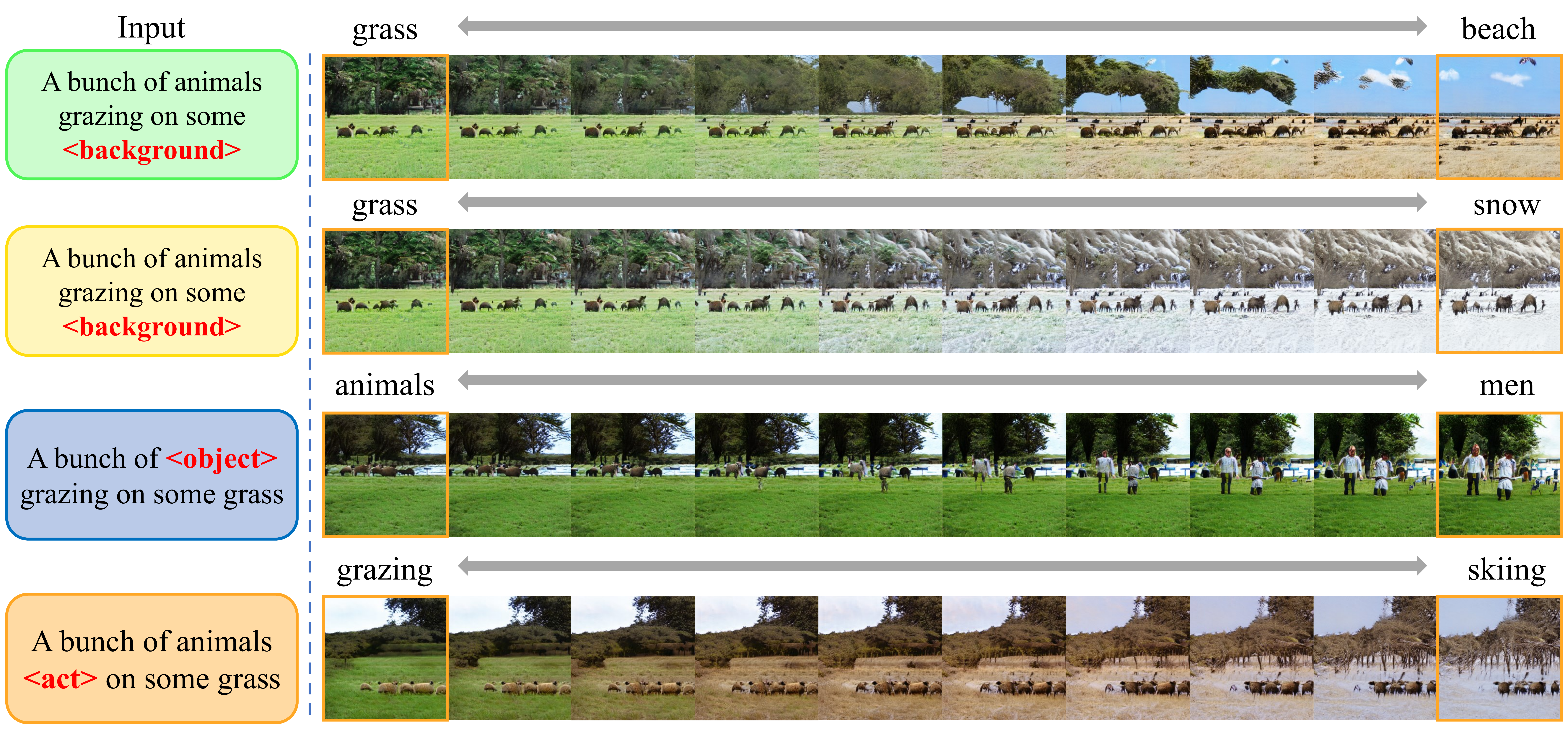}}
  \end{minipage}
  \caption{`Linguistic' interpolation of DiverGAN random latent-code samples on the COCO dataset, for four text input probes.}
  \vspace{-0.1in}
  \label{fig13} 
\end{figure}
\begin{figure}[t]
  \begin{minipage}[b]{1.0\linewidth}
  \centerline{\includegraphics[width=85mm]{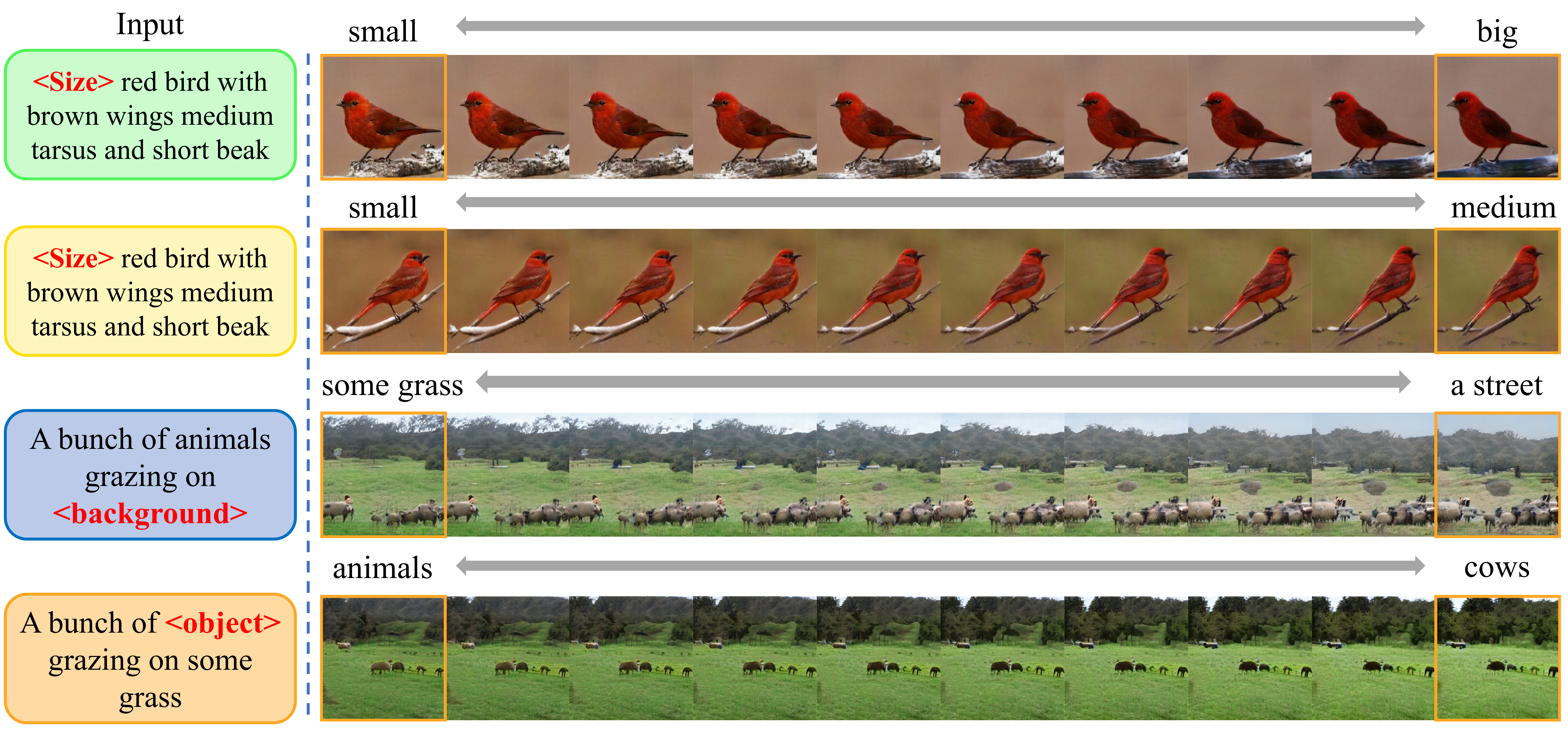}}
  \end{minipage}
  \caption{Unsuccessful `linguistic' interpolation of DiverGAN random latent-code samples on the CUB and COCO datasets, for four text input probes. For the third row, the desired attribute (i.e., a street) is not emerging.}
  \vspace{-0.1in}
  \label{fig14} 
\end{figure}
\subsubsection{Results of background flatten}
To prove the effectiveness of background flatten, we apply it to optimize the directions obtained by our proposed algorithm. The results are illustrated in Fig.~\ref{fig11}. By comparing the first row with the second row, we can see that the background is significantly improved and other image contents are maintained, indicating that the presented background-flattening method can be employed for existing latent-code manipulation approaches to fine-tune the backgrounds of synthetic samples. As can be observed in the third row, background flatten can also be leverage to remove the background while keeping the birds unchanged.  
\subsection{Results of a `linguistic' interpolation}
\subsubsection{Results of the linear interpolation between keywords}
\label{fc}
Fig.~\ref{fig12} shows the qualitative results of  the linear `linguistic’ interpolation of DiverGAN on the CUB bird data set, indicating that the attributes correlated with the synthesized sample do not always change gradually with the variations of word embeddings. For instance, the color of the bird does not vary continuously from `red’ to `blue’ in the first row. In the medium of interpolation results, DiverGAN generates multiple novel birds, whose bodies are composed of red and blue patches. However, the color attribute of the bird changes gradually from ‘red’ to ‘yellow’ in the second row. We are able to acquire an average color interpolation in RGB space by merging the first and second attributes.
We can also see that in the third row, the length of the beak varies smoothly along with textual vectors while other attributes remain unchanged. Furthermore, while the color of the beak changes continuously with the variations of word embeddings, the shape of the bird varies largely in the fourth row. The above results suggest that DiverGAN has the ability to capture the significant words (e.g., the color of the body and the length of the beak) in the given textual description. More importantly, by exploiting the characteristic as well as the linear interpolation between a pair of keywords, we can precisely control the image-generation process while producing various novel samples. 

The qualitative results of the linear interpolation between contrastive keywords on the COCO data set are shown in Fig.~\ref{fig13}. We can observe that DiverGAN accurately identifies `beach’, `snow’ and `men’ while generating the corresponding image samples. In addition, the background ($1^{st}$ and $2^{nd}$ row) and the object ($3^{rd}$ row) change continuously along with linguistic vectors.
It can also be seen that although we change the `acting' word from `grazing’ to `skiing’, the background significantly varies from `grass’ to `snow’ in the fourth row, which demonstrates that some words (e.g., `skiing’) play a vital role in the generation process of image samples. Furthermore, the above analysis indicates that when given adequate training images, DiverGAN is able to control the background (e.g., from grass to beach) and object (e.g., from animals to men) of complex scenes with the help of the linear `linguistic’ interpolation, since DiverGAN is able to learn the corresponding semantics in the linguistic space of the conditional input-text probes. 
\begin{figure}[t]
  \begin{minipage}[b]{1.0\linewidth}
  \centerline{\includegraphics[width=85mm]{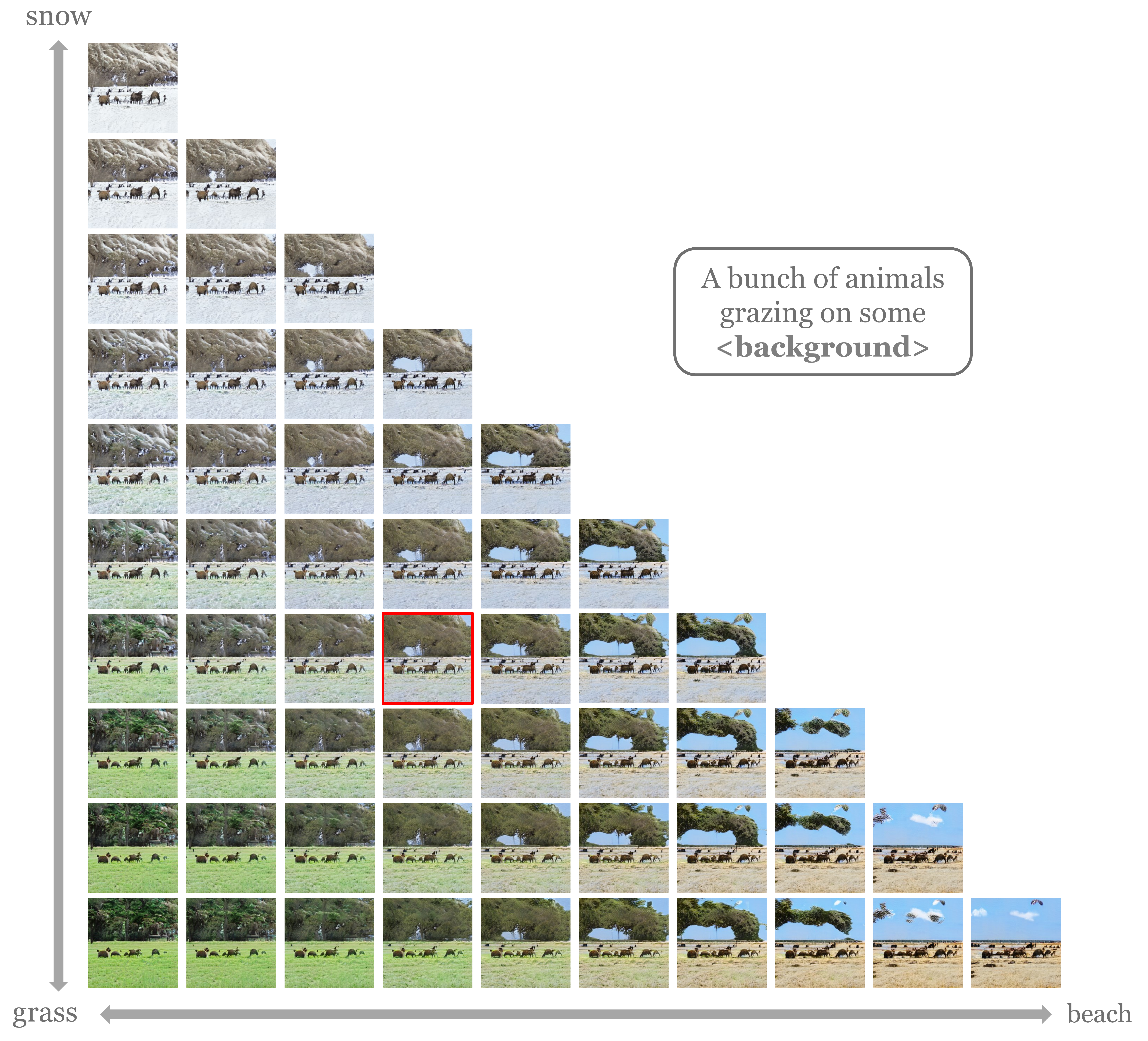}}
  \end{minipage}
  \caption{The triangular interpolation of latent codes, for linguistic attributes ${snow,grass,beach}$ on two dimensions. The center is marked in red.}
  \vspace{-0.1in}
  \label{fig15} 
\end{figure}

In addition to visualizing effective examples of the linear interpolation between keywords, we also present some unsuccessful results in Fig.~\ref{fig14}. As can be observed in Fig.~\ref{fig14}, the size of the bird ($1^{st}$ and $2^{nd}$ row) does not vary with the variations of the word (from `small’ to `big’ and from `small’ to `medium’). In addition, we can see that the background ($3^{rd}$ row) and the object ($4^{th}$ row) unfortunately do not change along with the word (from `grass’ to `street’ and from `animals’ to `cows’). At this point we can conclude that many meaningful contrasts can be learned (Fig.~\ref{fig13}), but there are areas where the method is not able to capture important variations along a dimension. This may be due to architectural or data-related limitations. In order to improve our insights, we will look at a triangular interpolation in the next subsection.

\subsubsection{Results of a triangular `linguistic' interpolation}
\label{4.3}
The triangular interpolation for linguistic attributes (i.e., the points between ${snow,grass,beach}$) in two dimensions is shown in Fig.~\ref{fig15}. We can observe that the transitions towards the three corner points are natural as well as smooth. Furthermore, the interpolation results achieve a balanced triangular shape within the triangle, such that the center marked in red is the combination of three linguistic attributes. If the application concerns data augmentation, 55 believable samples are obtained by performing the triangular interpolation between keywords.

\section{Conclusion}
\label{c}
In this paper, we propose several techniques to overcome the challenges of text-to-image generation in real-world applications. To ensure the quality of synthetic pictures, we created a {\em Good \& Bad} data set, both for a bird and a face-image collection, which comprises high-resolution as well as implausible synthesized samples, in which the images are chosen by following strict principles. Based on the {\em Good \& Bad} data set, we fine-tune the deep convolutional network trained on ImageNet to classify a generated image as ${Good}$ or ${Bad}$. To better understand and exploit the latent space of a conditional text-to-image GAN model, we introduce the independent component analysis (ICA) algorithm under an additional orthogonal constraint that can extract both independent and orthogonal components from the pre-trained weight matrix of the generator as the semantically-interpretable latent-space directions. In addition, we designed a background-flattening loss (BFL) to optimize the background appearance in the edited sample. To provide valuable insight into the relationship between the linguistic embeddings and the synthetic-sample semantic space, we conduct linear interpolation analysis between pairs of keywords. Meanwhile, we extend a pairwise linear interpolation to a triangular interpolation conditioned on three corners to further analyze the model. 

We evaluate our presented approaches on the recent DiverGAN generator that was pre-trained on three popular data sets, i.e., the CUB bird, Multi-Modal CelebA-HQ and MS COCO data sets. Extensive experimental results suggest that our well-trained classifier is able to accurately predict the quality classes of the samples from the testing set and our introduced algorithm can derive meaningful semantic properties in the latent space of DiverGAN, which validates the effectiveness of our proposed methods. 
Furthermore, we show that semantics contained in the image change gradually with the variations of latent codes, but the attributes of the sample do not always vary continuously along with the word embeddings. Moreover, we find that DiverGAN cannot capture the size of the object due to the mechanism of the convolutional neural network and cannot understand some words in the given textual description owing to the limitation of the data set.
In the future, we will explore how to utilize the presented approach to perform data augmentation for training image classifiers. Meanwhile, we plan to investigate the feasibility of adopting the proposed algorithm for the text-to-video generation task, which has various potential applications, such as synthesizing data for the reinforcement-learning system.

\bibliographystyle{model1-num-names}

\bibliography{cas-refs}




\end{document}